\documentclass[10pt,twocolumn,letterpaper]{article}

\usepackage[pagenumbers]{cvpr} 

\definecolor{cvprblue}{rgb}{0.21,0.49,0.74}
\usepackage[pagebackref,breaklinks,colorlinks,allcolors=cvprblue]{hyperref}

\makeatletter
\DeclareRobustCommand\onedot{\futurelet\@let@token\@onedot}
\def\@onedot{\ifx\@let@token.\else.\null\fi\xspace}

\def\eg{\emph{e.g}\onedot} 
\def\ie{\emph{i.e}\onedot}

\makeatother

\newcommand{\rcolor}{black}

\newcommand{\revision}[1]{\textcolor{\rcolor}{#1}}

\title{Generating Fit Check Videos with a Handheld Camera}

\author{Bowei Chen 
\qquad Brian Curless
\qquad Ira Kemelmacher-Shlizerman
\qquad Steve Seitz
\\
University of Washington\\
\\
{\tt\small \{boweiche, curless, kemelmi, seitz\}@cs.washington.edu}
}

\begin{document}
\twocolumn[{%
\renewcommand\twocolumn[1][]{#1}%
\maketitle
\begin{center}
    \centering
    \captionsetup{type=figure}
    \includegraphics[scale=1.3]{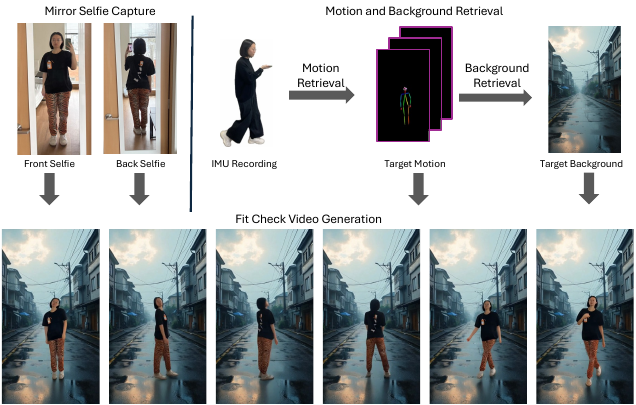}
    \vspace{-2mm}
        \captionof{figure}{         From two mirror selfie photos (top-left), we generate a photorealistic video of you performing a desired motion against a compatible desired background (bottom), with realistic shading, shadows, and reflections.  The motion is captured using your mobile phone's IMU sensors (top-right).  A target use case is self-captured ``fit check'' videos of you showing off an outfit.
        }
        \label{fig:teaser}
\end{center}%
}]

\begin{abstract}
Self-captured full-body videos are popular, but most deployments require mounted cameras, carefully-framed shots, and repeated practice. We propose a more convenient solution that enables full-body video capture using handheld mobile devices. Our approach takes as input two static photos (front and back) of you in a mirror, along with an IMU motion reference that you perform while holding your mobile phone, and synthesizes a realistic video of you performing a similar target motion. We enable rendering into a new scene, with consistent illumination and shadows. We propose a novel video diffusion-based model to achieve this. Specifically, we propose a parameter-free frame generation strategy and a multi-reference attention mechanism to effectively integrate appearance information from both the front and back selfies into the video diffusion model. Further, we introduce an image-based fine-tuning strategy to enhance frame sharpness and improve shadows and reflections generation for more realistic human-scene composition. 
\end{abstract}


\section{Introduction}
\label{sec:intro}

With the rise in popularity of video platforms like Tiktok and Instagram Reels, self-captured video has become a major industry.
However, taking high quality {\em full-body} videos of yourself is not straightforward. 
Anyone who has tried propping up their phone on a table for a selfie, knows how difficult it is to frame a shot this way.  For social media professionals, the most common solution is to use a tripod-mounted camera, but this introduces its own friction: creators must carry extra gear, guess at their framing without a reliable viewfinder, and often retake videos multiple times to achieve a desirable result~\cite{youtube_tripod_selfie,hertravelstyle_tripod_tips,peters2013tripod}. 

It would be much easier to stand in front of a full-length mirror and capture video with your mobile phone.  However, this introduces other problems: the camera is visible, the framing is awkward (Fig.~\ref{fig:teaser} top left), you have limited ability to move around, and the viewpoint moves with you.
These limitations highlight the need for a more convenient and user-friendly solution for full-body video self-capture.

Our goal is to enable full-body selfie video capture with a handheld mobile phone.  We employ a full-length mirror to capture front and back selfie photos, for a range of applications but with a particular focus on ``fit check'' videos (influencers trying on outfits) popular on social media.
Specifically, we provide the following capabilities:

\begin{itemize}
\item full-body self-capture with a handheld mobile phone that appears as if the camera was mounted on a tripod
\item video output generated from image captures (\ie, mirror stills) -- eliminating the need to record full-body video
\item accurate rendering of both the subject’s front and back
\item virtual backgrounds with consistent shading
\end{itemize}
In short, we seek to provide the video you would have achieved with a tripod-mounted camera and lots of practice in your desired environment.

Our solution is to separately capture your appearance (in a mirror), your movement (from phone sensors), and a compatible background, and then {\em compose} them together into a desired video (Fig. \ref{fig:teaser}).  All steps are designed to be easy and quick to achieve with a standard handheld mobile phone.


Solving this problem involves several challenges. 
First, we need to generate a video from two mirror stills (selfies) and motion information; the video should maintain temporal consistency, support human poses that differ from those in the input selfies, and accurately preserve the subject’s outfit and face -- all rendered from a stationary third-person viewpoint.
Second, the motion capture process must be simple and achievable using only a handheld phone -- this is especially challenging, as your body pose is high-dimensional and not fully observable from the phone that you are carrying in your hand.
Third, you should be composited into the new background with realistic shading, shadows, and reflections.

Our solution for motion self-capture is to leverage the inertial sensor (IMU) present on today's mobile phones.  Because your full body pose and motion are not typically fully observable from your handheld phone, we rely on a database of pose sequences and seek to retrieve the best match from IMU data.
Once a motion is selected, we retrieve the top-k backgrounds with the closest ground plane orientation to that of the selected motion’s background, ensuring motion–background compatibility (Fig. \ref{fig:teaser}, top right). 

\revision{Given the motion and background, a straightforward approach to generating fit check videos is to design a video diffusion-based human animation model that takes two mirror selfies, a target motion, and a background image as input.
However, adapting existing human animation methods~\cite{zhang2024mimicmotion,tu2024stableanimator} faces two challenges: (1) they do not support additional reference image inputs (\ie, back selfies), and simple modifications to include them lead to outfit inaccuracies; and (2) they often produce low-quality frames with blurriness and poor human-scene composition (\eg, weak reflections on reflective floors).
To overcome these issues, we introduce a novel, parameter-free frame generation strategy and a multi-reference attention mechanism that jointly enable effective use of all input selfies.
Furthermore, we propose an effective image-based fine-tuning strategy using a synthesized, task-specific image dataset to enhance sharpness, reflections, and shadow rendering while maintaining temporal consistency.}
Our contributions are as follows:
\begin{itemize}
\item We demonstrate a novel application to capture full-body, fixed-viewpoint fit-check videos from handheld mobile phones.  
Our approach supports customized motion capture and matching, and virtual background with reasonable shading and shadows.
\item We design a parameter-free frame generation strategy, as well as a multi-reference attention mechanism, to effectively integrate multiple reference images into human animation methods based on video diffusion models. 
\item 
\revision{
We introduce an image-based fine-tuning strategy that enhances frame quality, improving sharpness, reflections, and shadow rendering.
This architecture-agnostic strategy can be applied to other human animation models.
}
\end{itemize}




\section{Related Work}
\label{sec:related}



\noindent\textbf{Diffusion Models for Human Image Animation.}
Human image animation aims to animate a single static human image to generate a human video.
In this section, we review diffusion-based approaches, as they have shown superior generation quality compared to GAN-based methods~\cite{chan2019everybody,ren2020deep,siarohin2019animating,siarohin2019first,wang2020g3an,xu2022designing,yu2023bidirectionally,zhang2022exploring,zhao2022thin}. 


Diffusion-based approaches for human image animation can be categorized into two main types.
The first category~\cite{zhu2024champ, hu2023animateanyone, kim2024tcan, pang2024dreamdance, liu2024disentangling, xue2024follow, xu2023magicanimate, chang2023magicdance, hu2025animateanyone2,hu2025humangif,wang2025multi,wang2025anycharv}
 builds on top of an image diffusion model~\cite{rombach2022high}, where additional temporal layers, appearance and pose encoding strategy are designed based on the image diffusion model to ensure the temporal consistency, appearance and pose fidelity.  
For example, Animate Anyone~\cite{hu2023animateanyone} introduced ReferenceNet to inject reference image features into diffusion models and employed a pose guider (skeleton-based) to ensure accurate motion. Training occurred in two stages: first, diffusion model was fine-tuned on image pairs without temporal constraints; second, the added temporal layers (initialized by AnimateDiff~\cite{guo2023animatediff}) were optimized using multi-frame videos.
Champ~\cite{zhu2024champ} extended this approach by incorporating SMPL-based pose signals~\cite{SMPL-X:2019} for improved alignment. However, these methods suffer from jitter and temporal inconsistencies due to their reliance on image diffusion models, requiring extra training for temporal layers.

The second category~\cite{tu2024stableanimator, zhang2024mimicmotion, wang2024vividpose, tan2024animate, wang2024unianimate, shao2024360,lin2025omnihuman,wang2025unianimate,wang2025humandreamer,wang2025taming,zhao2025dynamictrl,ding2025mtvcrafter,zhou2025realisdance,wang2025unianimatedit}
 builds on video diffusion models~\cite{blattmann2023stable,yang2024cogvideox,wan2025} with stronger temporal consistency. 
StableAnimator~\cite{tu2024stableanimator} proposed an video diffusion-based pipeline with an identity-aware appearance controller to generate videos with improved identity preservation, conditioned on skeleton input. MimicMotion~\cite{zhang2024mimicmotion} introduced a confidence-aware pose guidance mechanism (in skeleton format) and regional loss amplification to enhance video quality and reduce distortions. However, these methods tend to produce frames that are slightly blurrier than image diffusion-based approaches.


In summary, both methods assume a single front-facing reference image without motion customization.
We build on MimicMotion for our task due to its strong appearance and temporal consistency and its wide adoption in the community~\cite{datou_mimic_motion,kijai_mimicmotion_workflow}.
We redesign the frame generation strategy and attention mechanism and introduce a fine-tuning stage, enabling the use of multiple reference images and producing sharper, higher-quality frames.
\revision{While based on MimicMotion, our pipeline can be readily adapted to other video diffusion-based methods with minimal changes.}

\noindent\textbf{Personalization Technique.}
This line of work personalizes foundation models for image~\cite{ruiz2023dreambooth,gal2022image,kumari2023multi,ruiz2024hyperdreambooth,ye2023ip,wang2024instantid,Li_2024_CVPR,he2024imagine} and video generation~\cite{wang2024customvideo,wei2024dreamvideo,polyak2024movie,he2024id,yuan2024identity,ku2024anyv2v,mou2024revideo}.
Personalized video synthesis methods generate videos from reference images, primarily using text prompts to control motion, background, and appearance.
In contrast, we enable finer control by incorporating a spatially aligned per-frame pose sequence and a specified background.

\noindent\textbf{Selfie-Related Applications.}
Numerous studies have studied selfies for tasks like reposing~\cite{ma2020unselfie}, face recognition~\cite{botezatu2022fun,kumarapu2023wsd}, style transfer~\cite{li2021adaptive}, novel view synthesis~\cite{athar2023flame,bian2021nvss,kania2022conerf,chen2024total}, but none address mirror selfies for full-body animation.

\section{Method}
\label{sec:method}

\begin{figure*}[!t]
    \centering
    \includegraphics[scale=0.85]{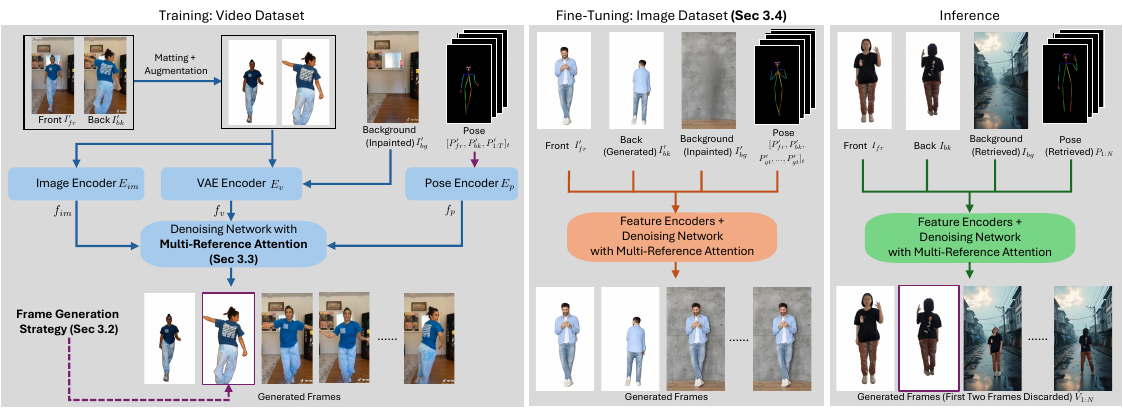}
        \vspace{-2mm}
    \caption{\textbf{Method Overview.} 
    \textit{Left}: We train our model on a fit-check video dataset using pairs of front and back images, GT video frames, an inpainted background, and a pose sequence.\protect\footnotemark{} Our frame generation strategy and multi-reference attention effectively encode features of multiple reference images.
    \textit{Middle}: We fine-tune the trained model on a high-quality image dataset, supervising the generated frames using the input front or back image to enhance frame quality.
    \textit{Right}: During inference, the method takes front and back selfies, a retrieved pose sequence, and a retrieved background as input, generating a video with the first two frames removed.
    The VAE decoders are omitted.
    }
    \vspace{-5mm}
    \label{fig:overview}
\end{figure*}

\noindent \textbf{Task Definition}. As input, the user captures two full-body mirror selfies -- front view $I_{fr}$ and back view $I_{bk}$ -- and records motion using a mobile phone's IMU. This  motion is used to retrieve a target pose sequence, $P_{1:N}$, where $N$ is the sequence length. Finally, a target background $I_{bg}$ is retrieved.
Given these inputs, we generate a fit check video, $\hat{V}_{1:N}$, with the person following the target poses within the target background (Fig.~\ref{fig:overview} right).  

\noindent \textbf{Training Data for Human Animation}.
\revision{
As no dataset contains paired mirror selfies and fit check videos, we build our dataset using fit check videos taken from third-person viewpoints. We found this data sufficient for training the model to handle selfie inputs effectively.
}
Each training pair consists of an input set $\{I'_{fr}, I'_{bk}, I'_{bg}, {P}'_{1:T}\}$ and its corresponding ground-truth (GT) video $V'_{1:T}$. Here, $V'_{1:T}$ comprises $T$ consecutive frames sampled from a fit check video $V'_{1:N}$ ($N\geq T$), while ${P}'_{1:T}$ represents the corresponding extracted poses in DWPose~\cite{yang2023effective} format. The background image $I'_{bg}$ is extracted under the assumption of a static camera, with occluded regions (not visible across all frames) inpainted using Stable Diffusion~\cite{rombach2022high}. The front and back view images, $I'_{fr}$ and $I'_{bk}$, are randomly sampled from $V'_{1:N}$ and have their backgrounds removed using ~\cite{ravi2024sam2}, as the original background is not needed during testing. 

Next, we first discuss the human animation component, assuming poses and background have been retrieved. We begin with a naive animation method and its limitations (Sec.~\ref{sec:naive}), which motivate our frame generation strategy (Sec.~\ref{sec:fg}) and multi-reference attention (Sec.~\ref{sec:mra}). Next, we introduce an image-based  fine-tuning strategy to enhance frame quality (Sec.~\ref{sec:fine-tuning}) and conclude with retrieval (Sec.~\ref{sec:retrieval}). Fig.~\ref{fig:overview} provides an overview of our method.

\vspace{-1.mm}
\subsection{Naive Method}
\label{sec:naive}

\revision{
Our naive method builds on MimicMotion~\cite{zhang2024mimicmotion}, a video-diffusion human animation framework that integrates pose and appearance features. The original model assumes one reference image and a driving pose sequence, which is incompatible with our setting. 
We thus extend its feature-integration mechanism to (1) aggregate appearance features from multiple reference images, and (2) condition generation on a target background image. 
We  will first describe the model training before detailing feature integration.
}

\noindent \textbf{Model Training}.
\revision{
We adopt the EDM diffusion framework~\cite{karras2022elucidating} for model training.
We use the ground-truth sequence $V=[I'_{fr}, V'_{1:T}]_t$ as supervision, where $[\cdot, \cdot]_t$ is temporal concatenation. Here, $I'_{fr}$ is treated as the first ground-truth frame. This follows MimicMotion's base model, an image-to-video method trained to generate the first frame identical to the input image.
We use $I'_{fr}$ instead of $I'_{bk}$ as MimicMotion assumes a front-facing input, and we continue training from their pretrained model.}
During training, we update the pose encoder (introduced later) and the denoising network. 
Please refer to the left box ``Training: Video Dataset'' in Fig.~\ref{fig:overview} for visualization. 
Note that, the naive method passes only the poses of $V=[I'_{fr}, V'_{1:T}]_t$ through the solid purple arrows, and does not generate the back image (bottom, purple border) in the output. 

\noindent \textbf{Feature Integration}. Three features are integrated into the denosing network. 
First, VAE feature $f_v = [E_v (I'_{fr}), E_v (I'_{bk}), E_v (I'_{bg})]_c$ is obtained by passing $I'_{fr}$, $I'_{bk}$, and $I'_{bg}$ through the VAE encoder $E_v$, where $[\cdot, \cdot]_c$ is channel-wise concatenation. 
This feature is duplicated along the temporal axis, channel-concatenated with the noisy latent, and then fed into the denoising network. 
Second, the image features are extracted as $f_{im} = E_{im}(I'_{fr}) + E_{im} (I'_{bk})$, where we use addition instead of concatenation, as it improves results in practice. Here, $E_{im}$ is implemented as a CLIP encoder~\cite{radford2021learning}, and its output feature $f_{im}$ is integrated into the denoising network via cross-attention layers. Third, pose features are extracted as $f_p = E_p([P'_{fr}, P'_{1:T}]_t)$, where $E_p$ is implemented the same as the pose encoder in  MimicMotion and $P'_{fr}$ is the DWPose  of $I'_{fr}$. Finally, $f_p$ is integrated into the denoising network by adding it to the output of denoising network's first layer. 
\revision{
All these modifications are natural extensions of design of MimicMotion.
}
\noindent \textbf{Reference Image Augmentation}. 
\revision{
During training, $I'_{fr}$ and $I'_{bk}$ are sampled from the same video as the GT sequence, resulting in similar lighting and shading. At test time, we use selfies and new backgrounds with different lighting.
}
To bridge this gap, we augment training data by adjusting the color tone of $I'_{fr}$ and $I'_{bk}$ to match random backgrounds using a pretrained image harmonization network~\cite{10285123} (details in supplementary). This approach outperforms image relighting-based augmentation~\cite{zhang2025scaling}, which tends to overmodify shading and distort clothing patterns.

\footnotetext{\vspace{-6pt} \noindent This example is for illustration only and is not used for training.}

\noindent \textbf{Model Inference}.
We replace $\{I'_{fr}, I'_{bk}, I'_{bg}, {P}'_{1:T}\}$ with $\{I_{fr}, I_{bk}, I_{bg}, {P}_{1:N}\}$ during inference.
Starting from a random noise, the model progressively denoises it to obtain a clean latent, which is then decoded by the VAE decoder to generate frames. We  remove the first frame because it is the same as $I_{fr}$.
Since this process only produces $T+1$ frames, we adopt the progressive latent fusion strategy from \cite{zhang2024mimicmotion} to generate $N$ frames ($N \geq T$). This method segments the sequence into overlapping segments with $T+1$ frames, denoises each segment, and fuses them to generate a N-length video $\hat{V}^{1:N}$.  
To better preserve identity, we refine the face in generated frames using a pretrained face refinement method~\cite{facefusion}, with $I_{fr}$ as the reference.
Fig.~\ref{fig:overview} (right) visualizes the process, where the naive method also does not produce the back-view frame with the purple border.

\begin{figure*}[!t]
\captionsetup[subfigure]{font=scriptsize,aboveskip=1pt}
\centering
\subcaptionbox{Inputs}%
{\includegraphics[width=0.129\linewidth]{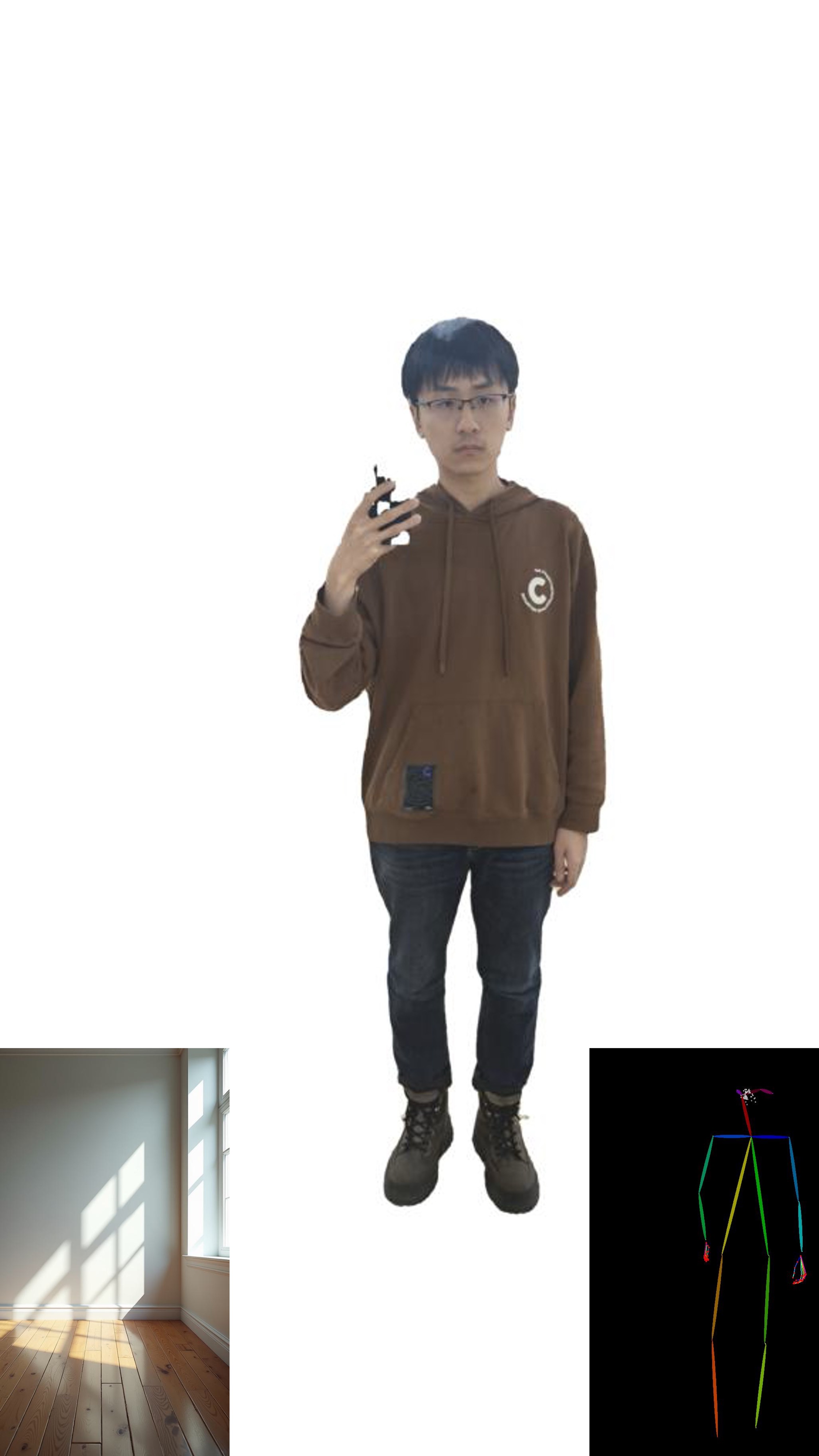}\hfill\includegraphics[width=0.129\linewidth]{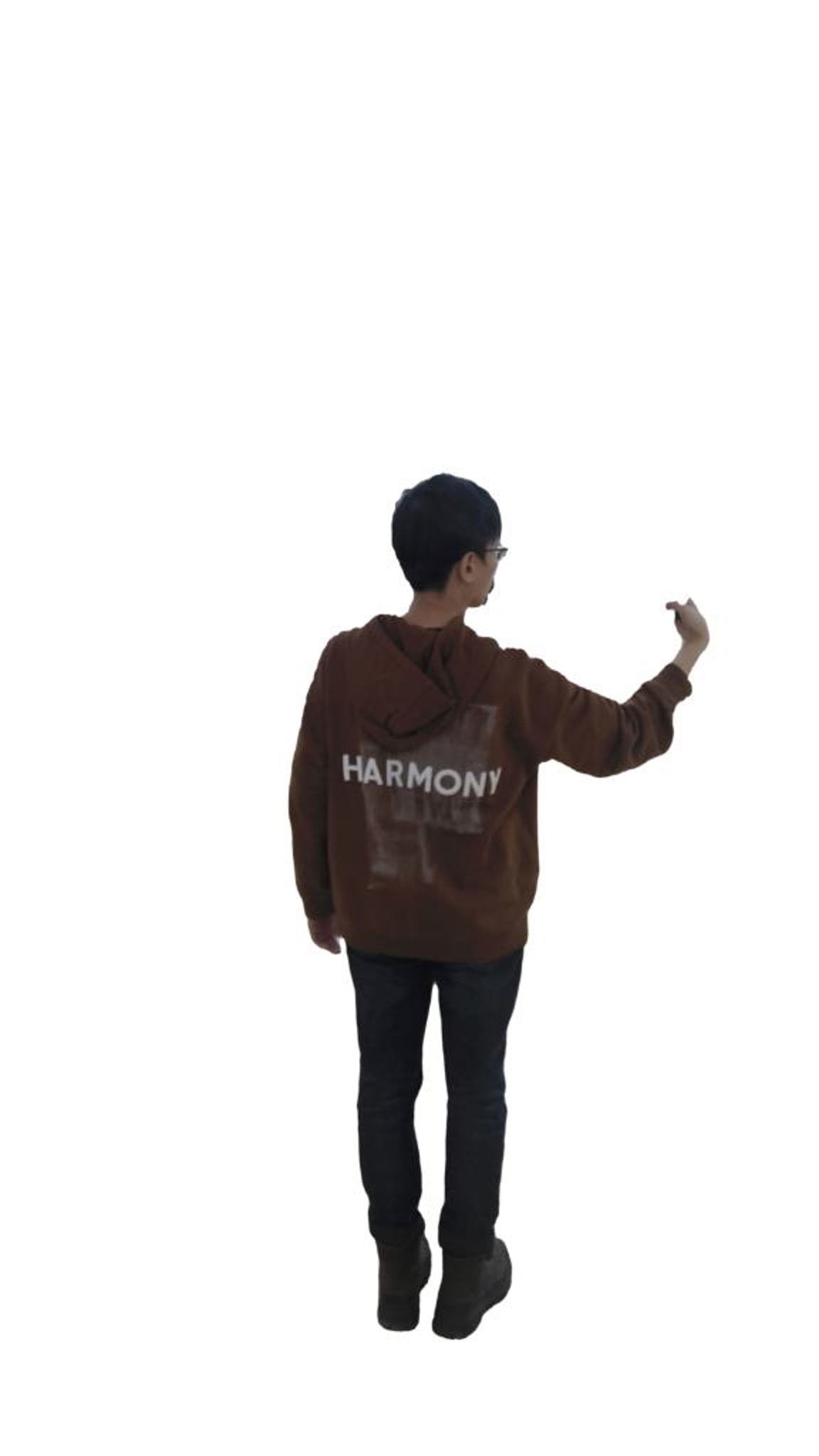}}
\subcaptionbox{Naive}%
{\includegraphics[width=0.129\linewidth]{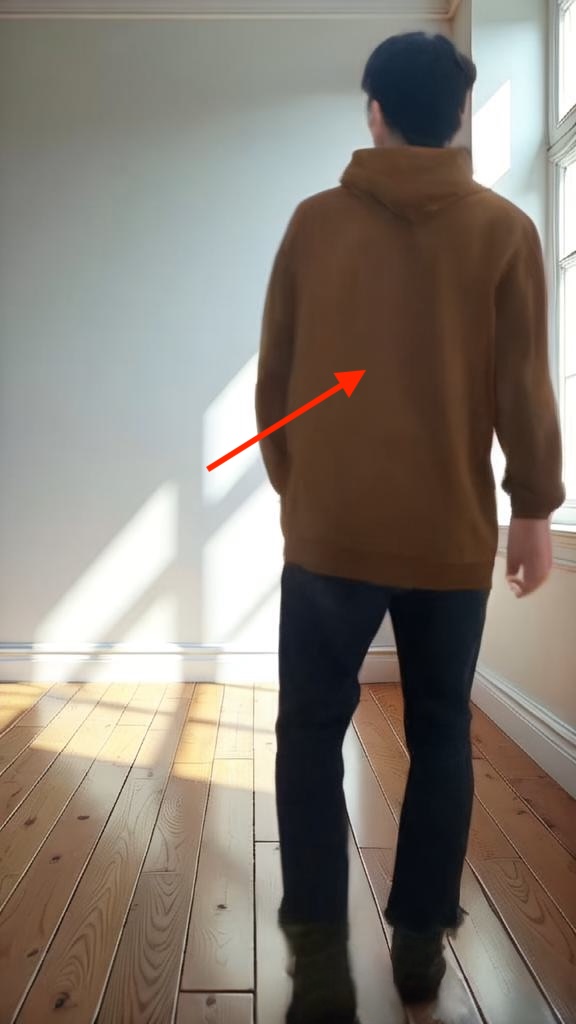}}
\subcaptionbox{Naive + RefNet}%
{\includegraphics[width=0.129\linewidth]{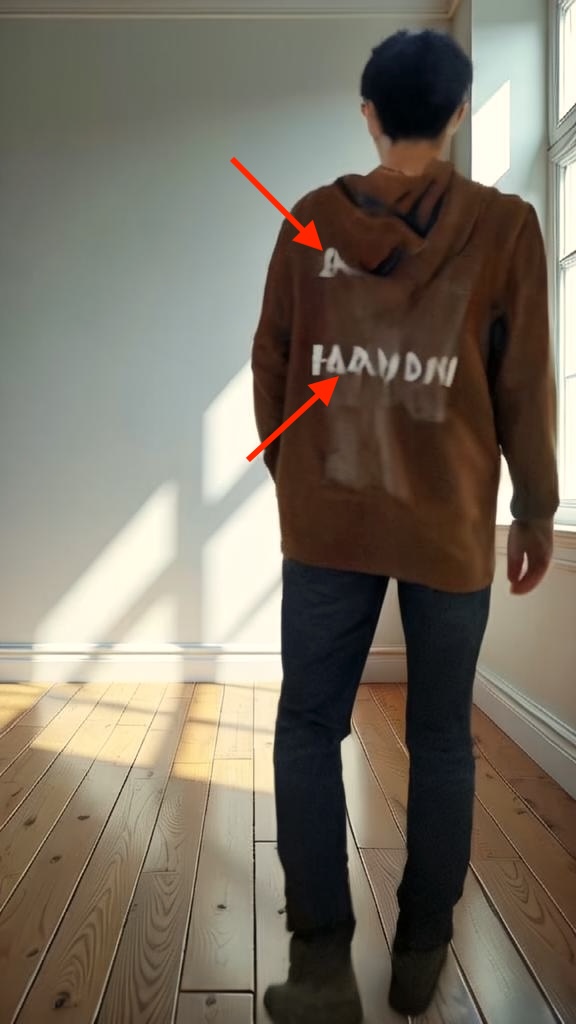}}
\subcaptionbox{Naive + FG}%
{\includegraphics[width=0.129\linewidth]{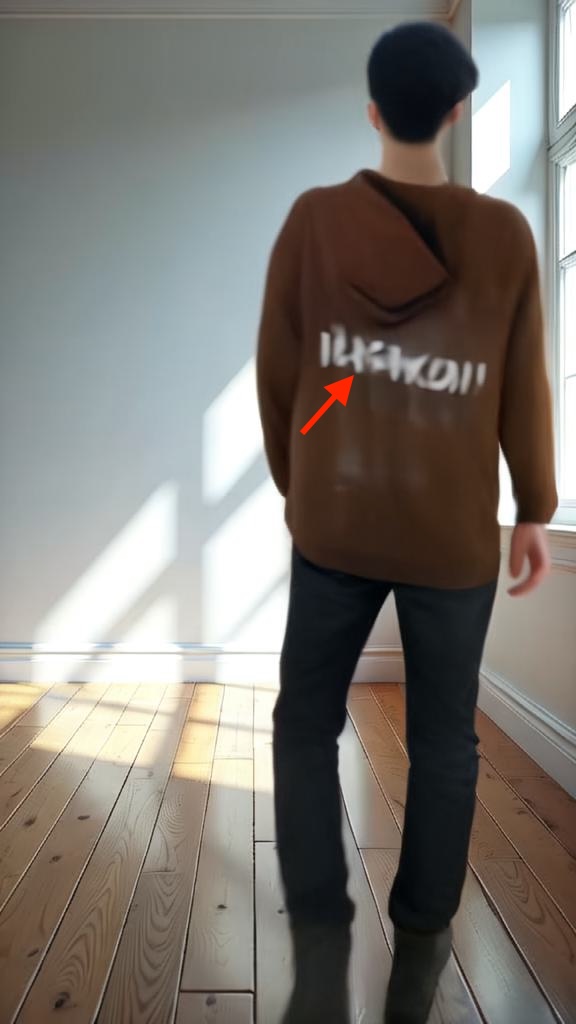}}
\subcaptionbox{Naive+FG+MRA}%
{\includegraphics[width=0.129\linewidth]{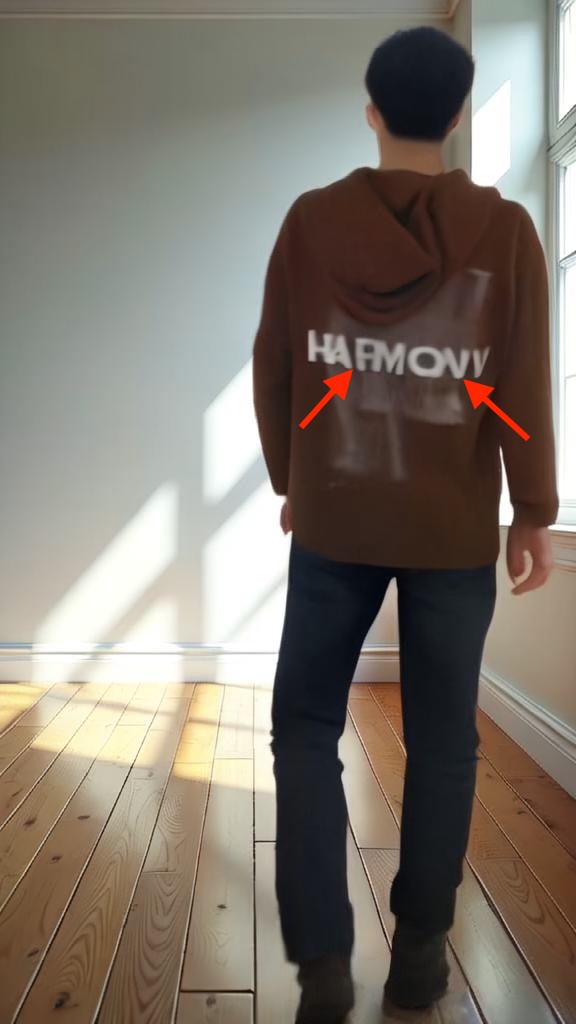}}
\subcaptionbox{Ours: (e) + FT}%
{\includegraphics[width=0.129\linewidth]{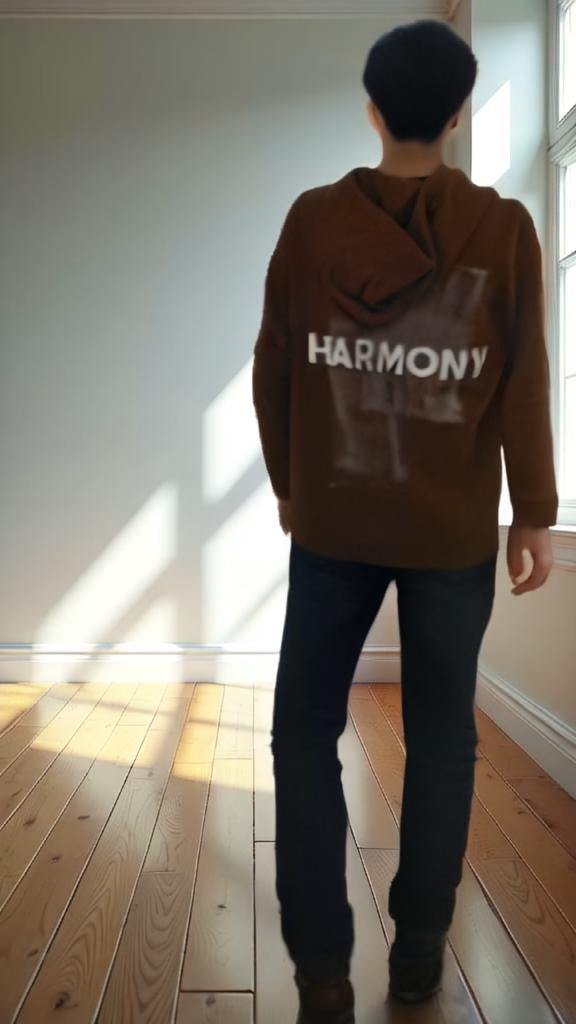}}
\vspace{-3mm}
\caption{\textbf{Model Ablations. } The inputs are shown in (a). The naive method (b) fails to render accurate back views. ReferenceNet (c) improves back-view generation but introduces hood artifacts and blurs text. Additionally, it requires extra parameters, reducing model efficiency. Our frame generation strategy (d) produces better back views than (b) without extra parameters, though text remains blurry. Multi-reference attention (e) enhances back view patterns, and adding the fine-tuning stage (f) delivers sharp, recognizable text. }
\label{fig:pipeline_ablation}
\vspace{-4mm}
\end{figure*}

Fig. \ref{fig:pipeline_ablation} (b) shows a frame generated by our naive design, which struggles to reconstruct patterns on the upper clothing in the back view. This shows that simply injecting features from feature encoders into the network is insufficient for effectively utilizing the additional reference image $I_{bk}$.

\vspace{-1mm}
\subsection{Frame Generation Strategy}
\label{sec:fg}
\revision{
The goal  is to improve the naive model by effectively integrating $I_{bk}$ for accurate back-view reconstruction. One option is to add a ReferenceNet~\cite{hu2023animateanyone,wang2024vividpose}, which extracts features from the front and back images and injects them into the denoising network via self-attention. As shown in Fig.~\ref{fig:pipeline_ablation}(c), this design captures back text (though blurry) but introduces unwanted white patterns around the hood. Moreover, ReferenceNet increases model parameters and training time by about 1.5×, making it computationally inefficient.
}

\revision{
We observe that the front-view clothing pattern is well reconstructed, likely because $I_{fr}$ is always generated as the first frame, allowing its features to propagate through the video due to strong frame-to-frame consistency.
}
Motivated by this, we propose a simple, parameter-free frame generation strategy (Fig.~\ref{fig:overview} left). Our insight is that enforcing the front and back views as the first frames leverages the video model’s consistency, ensuring key visual details, such as clothing patterns, remain coherent throughout the sequence.
Specifically, we treat the back-view image as the second ground-truth frame, enhancing the naive approach by setting $V = [I'_{fr}, I'_{bk}, V'_{1:T}]_t$ and $f_p = E_p([P'_{fr},  P'_{bk}, P'_{1:T}]_t)$, where $P'_{bk}$ represents the DWPose  of $I'_{bk}$. 
During inference, the model generates $T+2$ frames at once (through multiple denoising steps), and we discard the first two frames.

As shown in Fig.~\ref{fig:pipeline_ablation} (d), this substantially improves back-view accuracy over the naive method (b) and achieves results comparable to ReferenceNet (c), but with fewer parameters, reduced training time, and no hood artifacts. 
However, the text in (d) remains blurry and unreadable.

\vspace{-1mm}
\subsection{Multi-Reference Attention}
\label{sec:mra}

\begin{figure}[!t]
\centering
  \includegraphics[scale=0.9]{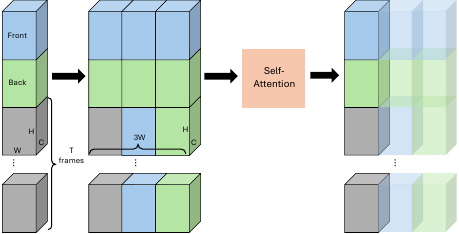}
  \vspace{-2mm}
\caption{\textbf{Multi-Reference Attention}. Given a pre-attention feature map (left), we duplicate and concatenate the front (blue) and back (green) features with all frame features (gray) (middle). For batch processing, the front and back features are also concatenated with themselves. The combined features then pass through self-attention layers, and we extract the first third of the output along the width axis as the final result (right).
}
\vspace{-5mm}
  \label{fig:mra}
\end{figure}

To better reproduce patterns from input images, we design a  multi-reference attention mechanism, building on top of our frame generation design.
Our key idea is to modify the denoising network’s spatial self-attention layers to better integrate features from multiple reference images into other frames (see Fig.~\ref{fig:mra}).
Specifically, we define a feature map from a self-attention layer as $x_{1:(T+2)} \in \mathbb{R}^{(T + 2) \times H \times W \times C}$, where H, W, and C represent the height, width, and channel size of the feature map, respectively. Based on our frame generation design, we extract the reference image features as $x_{fr} = x_{1}$ and $x_{bk} = x_{2}$, both having dimensions 
$1\times H \times W \times C$.  
Then we duplicate $x_{fr}$ and $x_{bk}$ along temporal axis T times. Next, we augment $x_{3:(T+2)}$ (\ie, feature maps of other frames, excluding $x_1$ and $x_2$) by concatenating them width-wise with $x_{fr}$ and $x_{bk}$.
Further, we concatenate  $x_{1}$ with $[x_{fr}, x_{fr}]_w$ and $x_{2}$ with $[x_{bk}, x_{bk}]_w$, both along the width dimension, for batch processing, where $[\cdot, \cdot]_w$ is width-wise concatenation. 
Finally, we apply self-attention and extract the first third of the output feature map along the width dimension as the output, similar to \cite{hu2023animateanyone}.
The key difference from \cite{hu2023animateanyone,wang2024vividpose} is that those methods apply self-attention using reference features from ReferenceNet, while ours directly uses features from the denoising network. Our design offers two advantages: (1) it ensures that $x_{fr}$, $x_{bk}$, and $x_{3:T+2}$ share the same feature space, enabling effective fusion, and (2) it is more efficient, avoiding extra parameters by removing the need for ReferenceNet.

As shown in Fig. \ref{fig:pipeline_ablation} (e), adding multi-reference attention sharpens text details compared to (d), though the word ``HARMONY'' remains a bit hard to recognize.


\subsection{Image-Based Fine-Tuning }

\label{sec:fine-tuning}
To improve frame quality, we introduce an image-based fine-tuning strategy consisting of two steps:
(1) constructing a task-specific, high-quality image dataset, and
(2) fine-tuning the model on this dataset while maintaining temporal coherence.
\revision{
We did not apply high-quality video fine-tuning due to two challenges:
(1) it is difficult to obtain fit check videos (with both front and back views) of higher quality than our training data, and
(2) motion blur in videos often degrades sharpness. 
Fig.~\ref{fig:overview} (middle) shows a visualization. 
}

\noindent \textbf{Constructing Image Dataset for Fine-Tuning}. 
Our goal is to construct training pairs, each consisting of an input set $\{I'_{fr}, I'_{bk}, I'_{bg}, P'_{gt}\}$ and a corresponding ground-truth image $I'_{gt}$. Obtaining real-world images $I'_{fr}$, $I'_{bk}$, and $I'_{gt}$ of the same person within identical backgrounds is challenging; thus, we opt for synthetic dataset creation. 
\revision{
To construct each training pair, we first collect real front-view images ($I'_{fr}$) from the web, prioritizing those with visible reflections or shadows to improve the model’s ability to render realistic lighting effects. }
Next, we employ a pretrained image reposing method (the first stage of \cite{hu2023animateanyone}) to synthesize the corresponding back-view images ($I'_{bk}$).
We set the ground-truth $I'_{gt}$ as either $I'_{fr}$ or $I'_{bk}$, obtaining $I'_{bg}$ by inpainting the foreground and shadow or reflection regions of $I'_{gt}$.
${P}'_{gt}$ is derived from the selected $I'_{gt}$.
 In addition,  $I'_{fr}$ and  $I'_{bk}$ undergo matting when used as input.

\noindent \textbf{Fine-Tuning Video Models on Image Dataset}.  
We fine-tune our trained model using constructed image pairs, where the ground-truth image $I'_{gt}$ and its corresponding pose $P'_{gt}$ are duplicated $T$ times to match the required input sequence length.
In practice, this duplication strategy leads to better temporal consistency compared to fine-tuning with non-duplicated sequences (\ie, setting $T = 1$).
Inspired by~\cite{chen2024videocrafter2}, we fine-tune only the spatial layers of the model to preserve its temporal coherence.
During fine-tuning, we select $I'_{fr}$ as GT 80\% of the time, and $I'_{bk}$ 20\% of times. This balance is chosen because $I'_{bk}$ is synthetically generated and may contain artifacts, while completely omitting it would lead to model collapse, causing the model to generate only front views, even for back-facing input poses.
As part of data augmentation, we randomly shift and scale $I'_{gt}$, $P'_{gt}$ and $I'_{bg}$, filling undefined regions with white pixels.


Fine-tuning on high-quality images with shadows and reflections for just 30 minutes on a single NVIDIA A100 GPU improves frame sharpness and human-scene composition (\ie, more realistic shadows and reflections). As shown in Fig.~\ref{fig:pipeline_ablation} (f), it also sharpens text. 
Our fine-tuning strategy is architecture-agnostic and applicable to other human animation models.
See supplementary for examples of shadow and reflection improvements.


\vspace{-1mm}
\subsection{Motion and Background Retrieval from IMU}
\label{sec:retrieval}
\noindent\textbf{Motion Acquisition}.
A motion generation model could map IMU data to motion, but existing approaches have key drawbacks: (1) they require multiple IMUs or motion capture devices~\cite{van2024diffusionposer,PIPCVPR2022,du2023avatars,jiang2022transformer,yuan2023physdiff}, while we use only a single phone, or (2) they suffer from foot sliding and jittering with sparse input~\cite{mollyn2023imuposer,xu2024mobileposer,tevet2023human,zhang2023generating}.
Given these limitations, we adopt motion retrieval instead.
However, prior retrieval methods~\cite{petrovich23tmr,bica2024improving,Fujiwara_2024_ECCV,shashank2024morag,li2024lamp} rely mainly on text queries, which do not suit our setting.
Training an IMU-to-motion retrieval model is also challenging due to the lack of large-scale paired mobile IMU and motion datasets, and synthetic data~\cite{AMASS:ICCV:2019} introduce a domain gap, as real phone IMUs are noisy and optimized for non-motion tasks.

Instead, we propose an IMU-based motion retrieval approach that matches recorded orientation and translation to a database of fit check motions.
The idea is to retrieve the top-k closest matches by computing dynamic time warping (DTW) distances between the recorded and candidate motions' orientation and translation (details in supplementary).



\noindent\textbf{Background Acquisition}.
The retrieved motion must align naturally with the virtual background’s ground plane.
Given the retrieved motion, we select the top-k backgrounds whose ground plane orientation best matches that of the motion’s original background (details in supplementary).



\vspace{-2mm}
\section{Experiments}
\vspace{-2mm}
In this section, we evaluate the human animation component.
Implementation details and motion retrieval evaluation are provided in the supplementary.

\noindent\textbf{Datasets.}
\label{sec:dataset}
We collect 1,590 fit check videos from the web, each featuring a single person with both front and back views. The average length is 9.5 seconds, and subsampling every 4 frames yields about 72 frames per video. The dataset is split into 1,441 training and 149 test videos.
Body orientation is estimated per frame using a pretrained SMPL-based detector~\cite{li2022cliff}, classifying frames as front (330°–30°) or back (150°–210°).
For fine-tuning, we collect 122 front-view web images to form 122 fine-tuning pairs.
In addition to our test set, we also evaluate on widely used datasets, TikTok~\cite{Jafarian_2021_CVPR_TikTok_dataset} and UBC Fashion~\cite{zablotskaia2019dubc_dataset}.

To evaluate our selfie task, we create a self-capture dataset with front and back mirror selfies from 8 individuals wearing various outfits, totaling 24 captures. 
We record a ``ground truth'' fit check video for each capture, maintaining the same clothing and a nearly identical background image.





\begin{figure*}[!t]
    \centering
\includegraphics[scale=0.63]{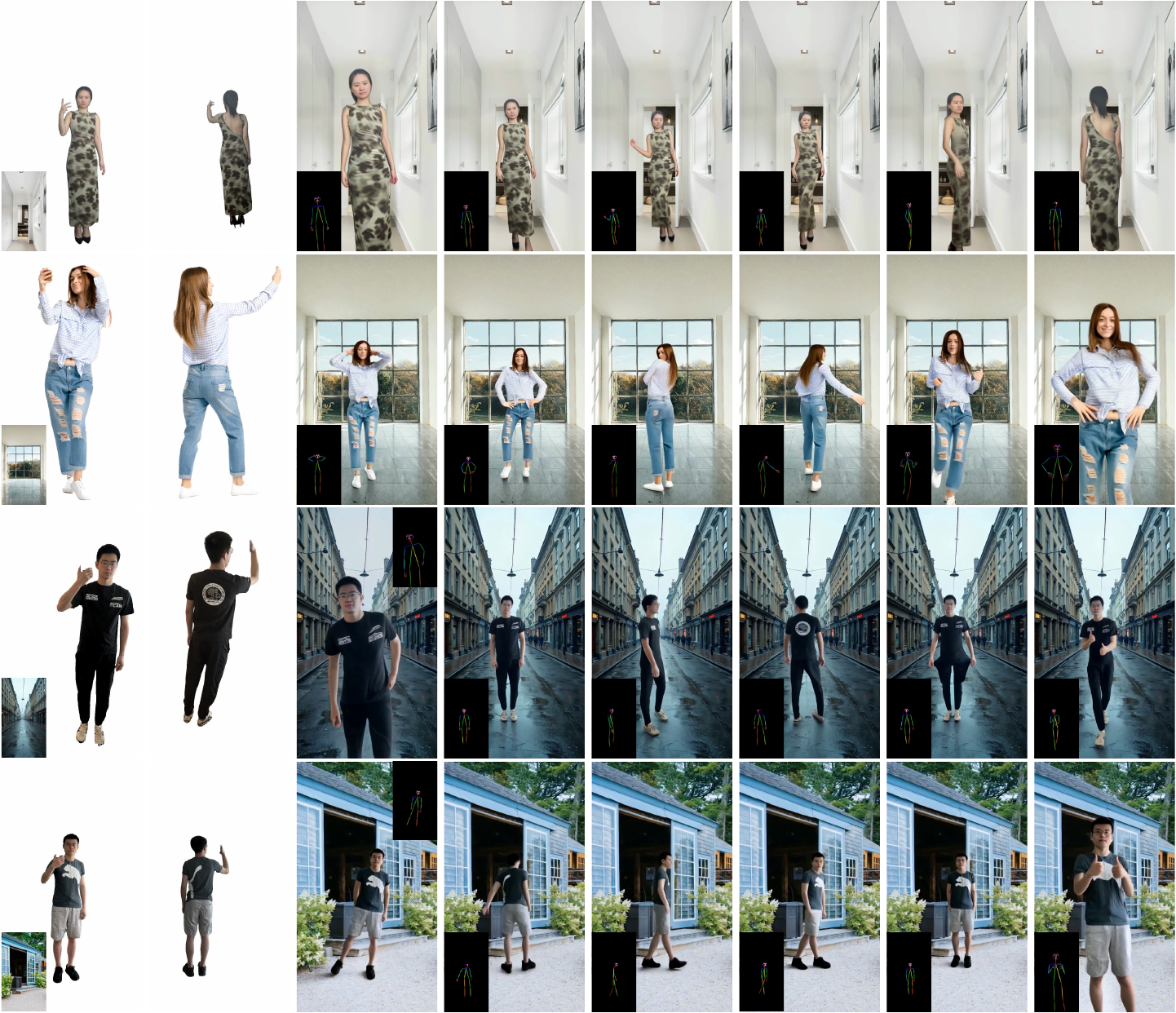}      \vspace{-3mm}
    \caption{\textbf{Our Results.} The left two columns show the input selfies and background, while the right six columns display the generated results (inset: pose input, locations adjusted to avoid occlusion). Given mirror selfies with various outfits and lighting conditions, our method generates realistic fit-check videos, accurately capturing appearance across diverse poses. Additionally, it generates reflections (rows 1–3) and shadows (row 4) on the ground, ensuring natural integration with both indoor and outdoor backgrounds.
    }
    \vspace{-4mm}
    \label{fig:main_results}
\end{figure*}

\begin{figure}[!t]
    \centering
\includegraphics[scale=0.34]{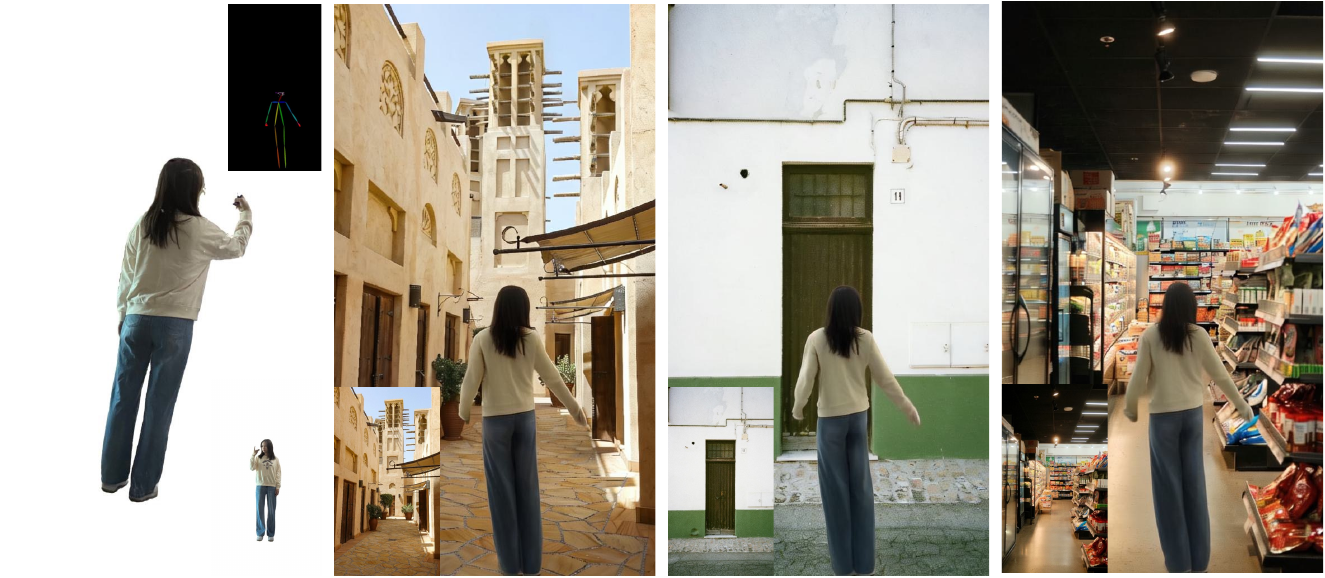}  
    \vspace{-2mm}
    \caption{\textbf{Results with the Same Capture under Different Virtual Backgrounds.} The first column shows the input selfies and target pose; the others show results under different virtual backgrounds (insets: input backgrounds). Despite strong left lighting in the selfies, our method adapts shading to each background.
    }
    \vspace{-4mm}
    \label{fig:diff_bg}
\end{figure}
\begin{figure*}[!t]
  \centering



  \subcaptionbox{Inputs}[.133\linewidth]{%
    \includegraphics[width=\linewidth]{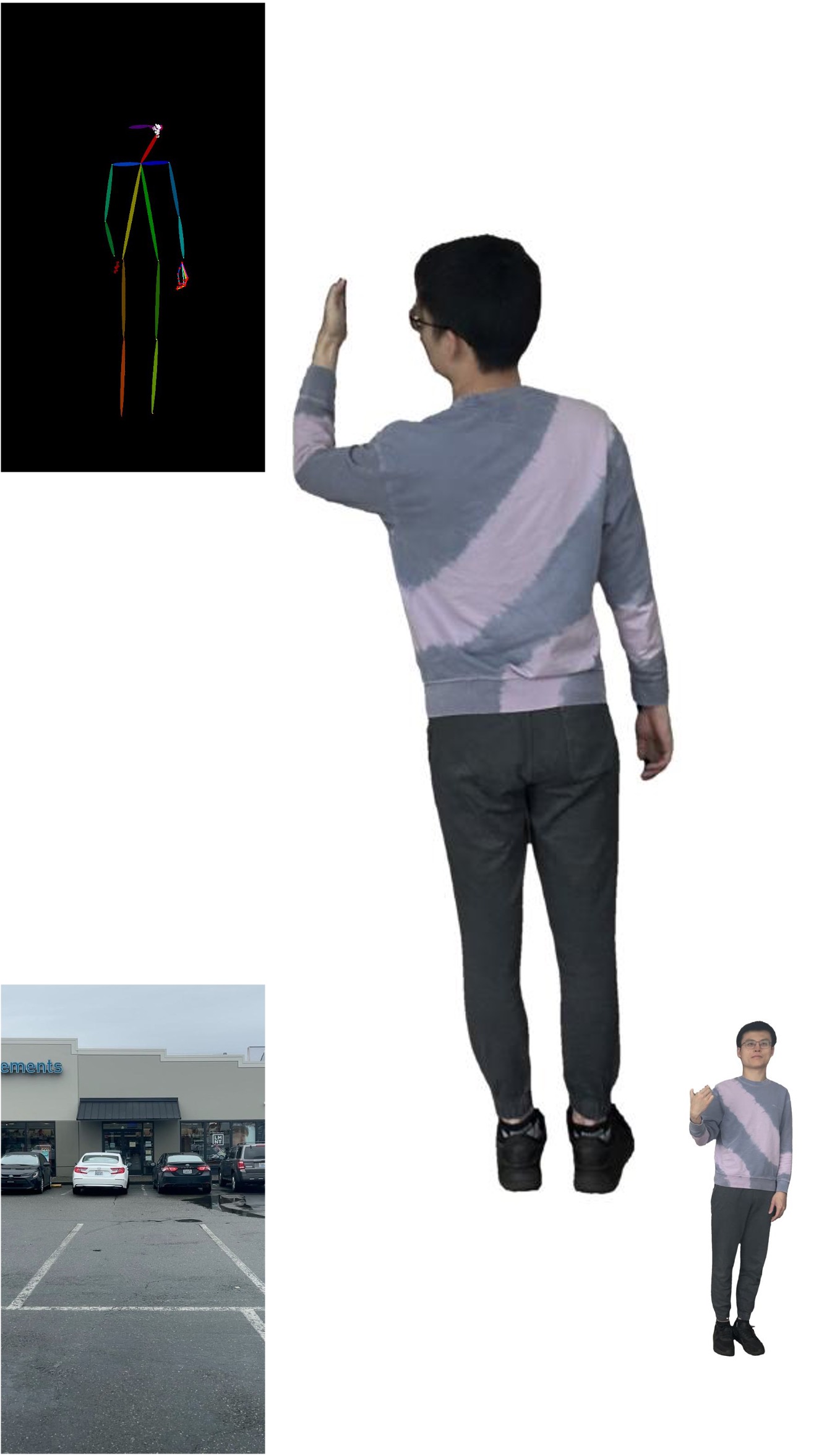}}
  \subcaptionbox{Animate Anyone}[.133\linewidth]{%
    \includegraphics[width=\linewidth]{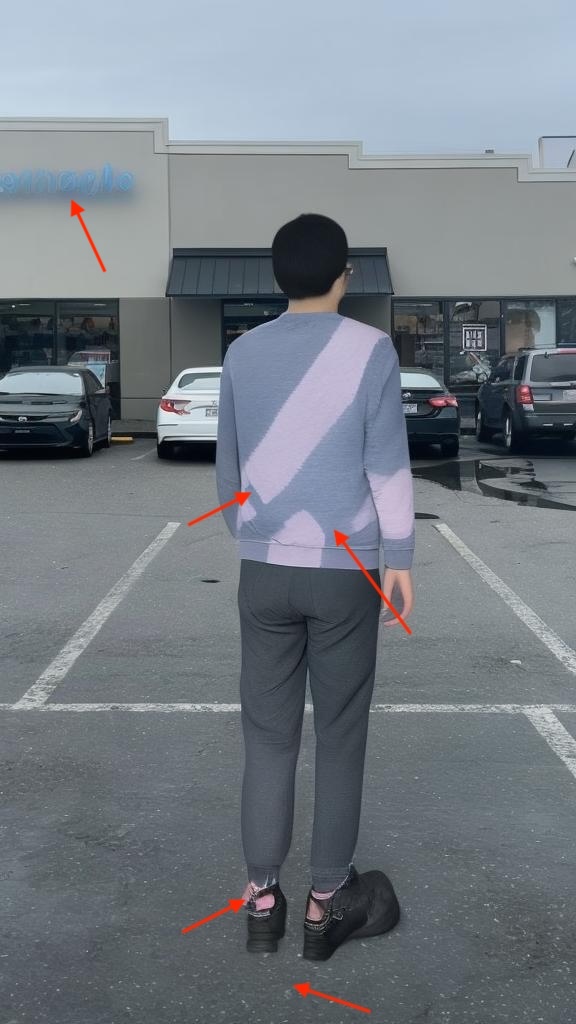}}
  \subcaptionbox{Champ}[.133\linewidth]{%
    \includegraphics[width=\linewidth]{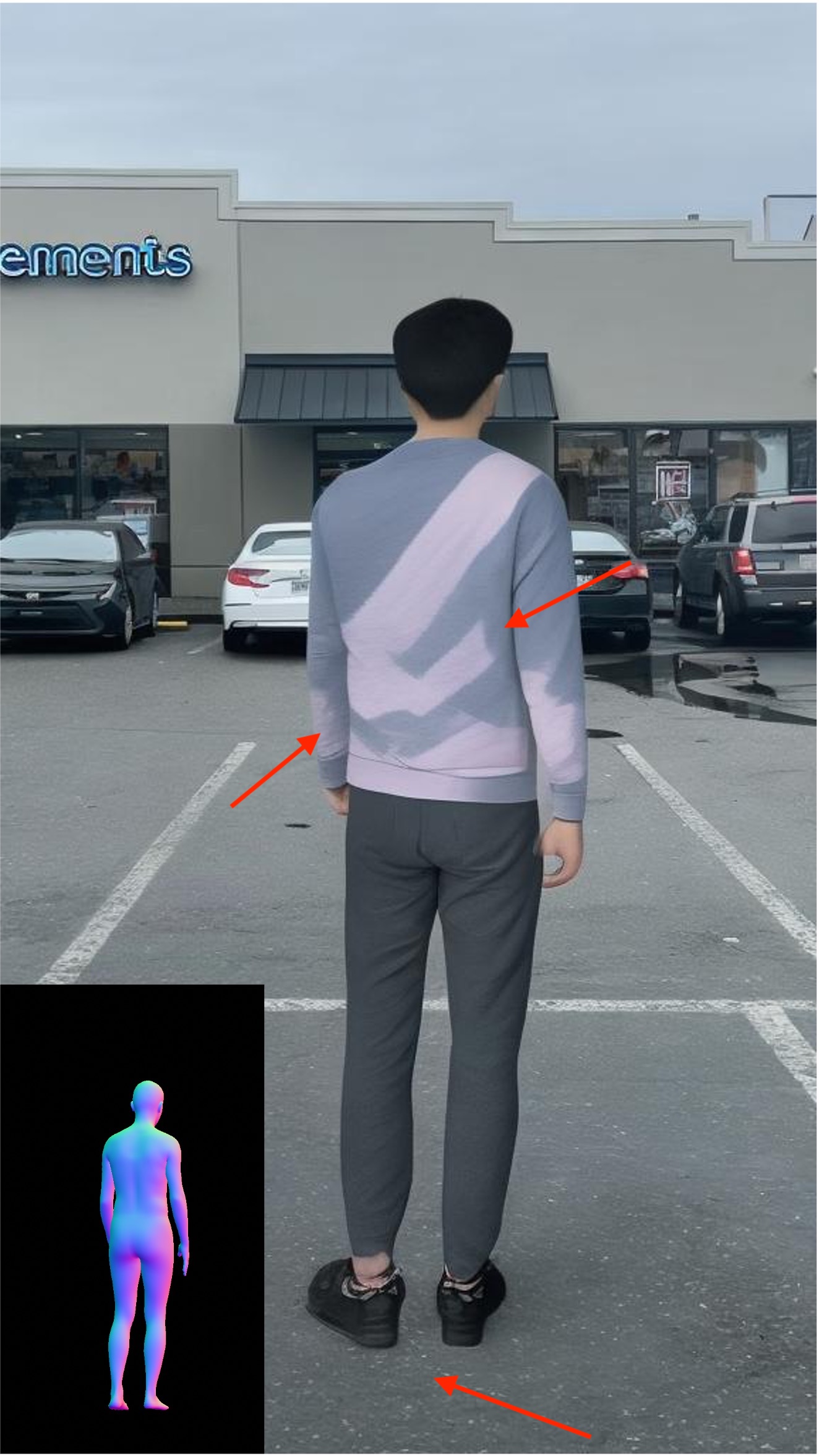}}
  \subcaptionbox{StableAnimator}[.133\linewidth]{%
    \includegraphics[width=\linewidth]{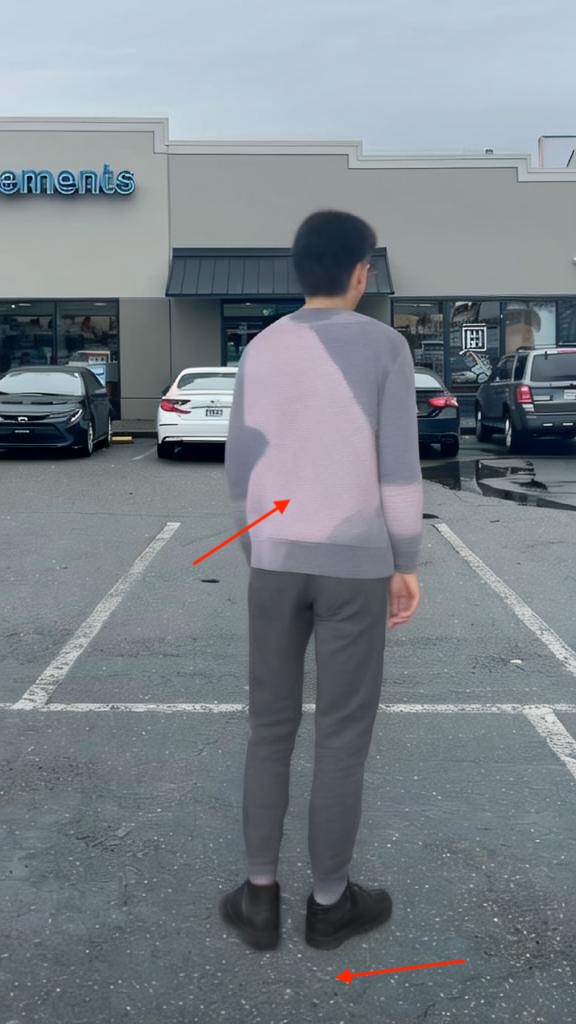}}
  \subcaptionbox{MimicMotion}[.133\linewidth]{%
    \includegraphics[width=\linewidth]{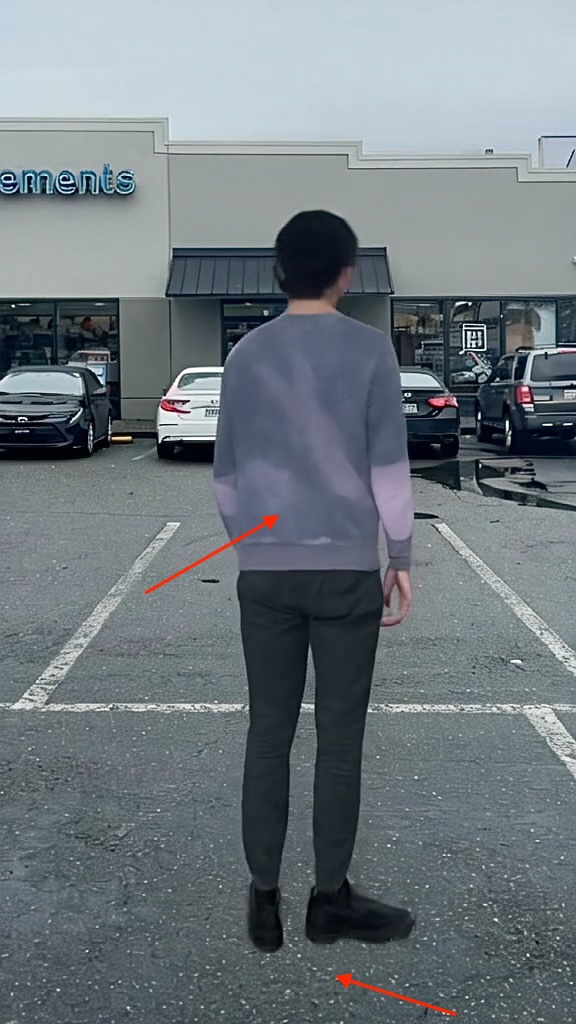}}
  \subcaptionbox{Ours}[.133\linewidth]{%
    \includegraphics[width=\linewidth]{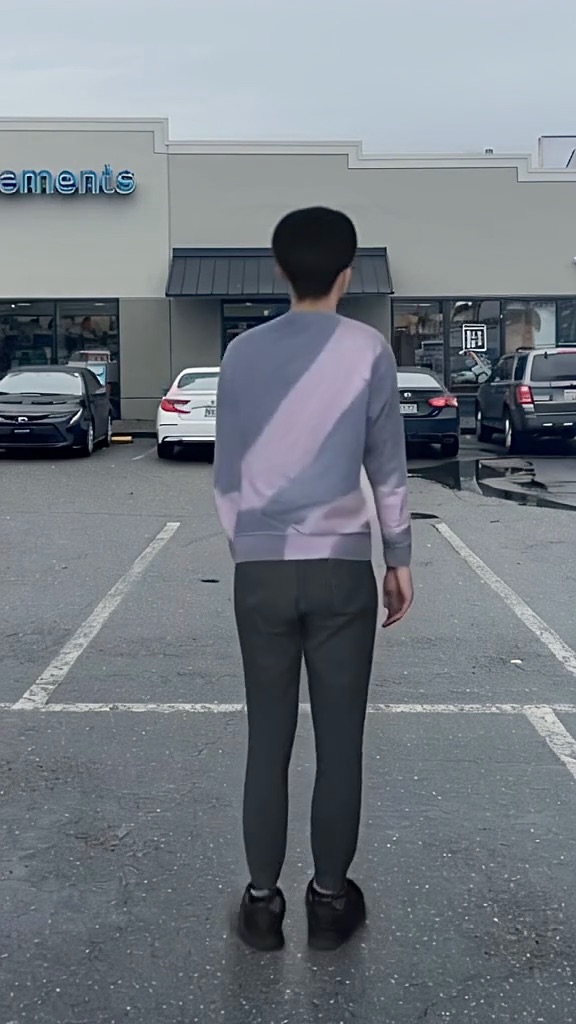}}
  \subcaptionbox{Real Frame}[.133\linewidth]{%
    \includegraphics[width=\linewidth]{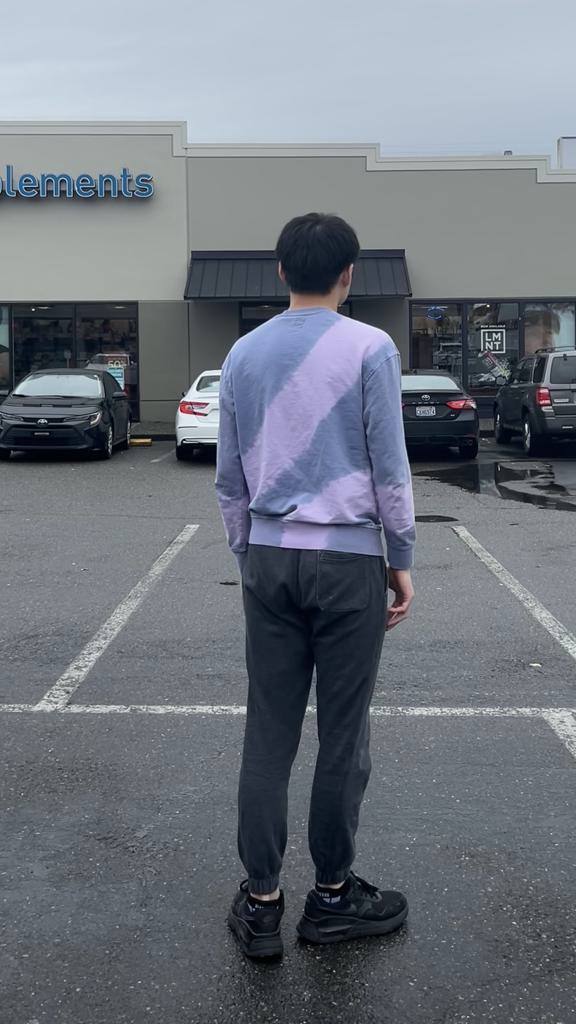}}
  \vspace{-2mm}
  \caption{
    \textbf{Qualitative comparison with baselines.}
    Inputs and real frame are shown in (a) and (g), respectively. The input pose is detected from the real frame. Image diffusion-based methods (b) and (c) fail to generate accurate back views and exhibit poor temporal consistency, causing background jitter and artifacts (see top-left blue text in the first row of (b)). Video diffusion-based baselines (d) and (e) generate completely wrong back views.
    Our method (f) achieves accurate outfit rendering and realistic ground reflections.
    Note that background images and real videos, though captured in the same session, may vary in intensity and color tone due to auto-exposure and white balance.
  }
    \vspace{-4mm}
  \label{fig:baseline_comparison}
\end{figure*}

\noindent\textbf{Our Results.}
We present the results of our method in Fig.~\ref{fig:main_results}.
First, our method generates realistic fit-check videos with diverse poses while consistently preserving identity, even from selfie inputs.
Second, our method effectively leverages both the front and back selfies to reconstruct a plausible outfit appearance across various clothing types (\eg, dresses, pants, shorts). Observe the detailed rendering of the person holding the pants in row 3, column 7.
Third, our method naturally composites the person into diverse indoor and outdoor backgrounds with realistic lighting, reflections, and shadows. Please see more results in supplementary.

Fig.~\ref{fig:diff_bg} shows results of the same capture composited into various virtual backgrounds.
Despite the input selfie having strong left-sided lighting, our method successfully relights the person to match each background’s illumination, enabled by our reference image augmentation strategy and the learned prior of the video diffusion model.

\noindent\textbf{Baselines.}
To our knowledge, no prior work directly addresses our task. So we adapt four human animation methods as baselines:
 (1) Animate Anyone~\cite{hu2023animateanyone} uses ReferenceNet to extract features from a single reference image. We extend it to extract features from both front and back selfies and integrate them into the denoising network. The background image is encoded by the VAE and incorporated into the denoising network in the same manner as noisy latents, aligning with our design.
(2) Champ~\cite{zhu2024champ} builds on Animate Anyone with additional pose signals (\eg, SMPL). We adapt it in the same way as (1) with additional pose inputs.
(3) StableAnimator~\cite{tu2024stableanimator}.
(4) MimicMotion~\cite{zhang2024mimicmotion}.
Both (3) and (4) are video diffusion-based methods adapted in the same way as our naive method, except without reference image augmentation.
All baselines share our input setup (except Champ with additional pose input). For fair comparison, we initialize each model with its pretrained checkpoint and continuing training them on our dataset. More details are in supplementary.


\noindent\textbf{Comparison with Baselines.}
First, we present a qualitative comparison with baselines on our selfie dataset in Fig.~\ref{fig:baseline_comparison}. 
Image diffusion-based methods, Animate Anyone (b) and Champ (c), leverage back selfies via ReferenceNet to improve back-view generation but introduce artifacts and temporal inconsistency, causing background jitter.
All baselines fail to produce realistic ground reflections, reducing realism, while our method generates reasonable reflections, demonstrating the effectiveness of the shadow and reflection enhancement in the fine-tuning stage.
Overall, our method delivers accurate outfits, realistic reflections, and strong temporal consistency.


Tab.~\ref{tab:main_quantitative} presents the quantitative results. Champ achieves a better LPIPS score than other baselines due to its sharp image diffusion-based frame generation. However, both Champ and Animate Anyone perform poorly on video metrics (FID-VID, FVD), indicating temporal inconsistency. Our model outperforms all baselines across all metrics. Additional qualitative and quantitative comparisons on selfie captures, datasets (our test set, UBC Fashion, TikTok), and videos are in the supplementary.

\noindent\textbf{Human Study.} \revision{We conducted a user study with 21 participants, each rating four identical videos (on a 1–5 scale, higher is better) across three aspects: \textit{body size accuracy}, \textit{garment accuracy}, and \textit{realism}. Our method achieved average scores of 3.88, 4.08, and 3.75, outperforming the best baselines -- 3.20 (MimicMotion), 2.88 (Champ), and 2.82 (MimicMotion), respectively. See supplementary for more quantitative results for the three aspects. 
}
\begin{table}[!t]
\centering
\caption{\textbf{Quantitative Comparisons on Self-Captured Dataset.} All methods are evaluated without face refinement as post-processing and the target poses are detected from the captured real videos. Our method outperforms all  baselines and variants.}
\vspace{-3mm}
\resizebox{0.47\textwidth}{!}{
\begin{tabular}{|l|cccccc|}
\toprule
       {Method}  &  {LPIPS} $\downarrow$       & {SSIM} $\uparrow$  & {PSNR} $\uparrow$   & {FID} $\downarrow$   & {FID-VID} $\downarrow$   & {FVD} $\downarrow$   \\
\midrule
Animate Anyone~\cite{hu2023animateanyone}       & 0.428  & 0.429  & 14.12  & 128.9   & 65.33  & 1158 \\
Champ~\cite{zhu2024champ}      & 0.410  & 0.402  & 13.86  & 121.8   & 61.17  & 1108 \\
StableAnimator~\cite{tu2024stableanimator} & 0.422  & 0.454  & 14.70  & 131.5   & 60.09  & 982.8 \\
MimicMotion~\cite{zhang2024mimicmotion}      & 0.413  & 0.458  & 14.69  & 123.6   & 59.63  & 947.8 \\
\midrule
Ours-FG-MRA-FT   & 0.411  & 0.459  & 14.67  & 123.5   & 60.41  & 925.4 \\
Naive + RefNet     & 0.403  & 0.463  & 14.77  & 123.4   & 59.58  & 922.2 \\
Ours-MRA-FT     & 0.398  & 0.468  & 14.98  & 123.4   & 57.00  & 921.6 \\
Ours-FG     & 0.404  & 0.462  & 14.92  & 124.2   & 53.45  & 895.8 \\
Ours-MRA      & 0.386  & 0.478  & 16.17  & 118.6   & 53.92  & 871.0 \\
Ours-RIA      & 0.384  & 0.491  & 16.36  & 119.0   & 52.98  & 866.6 \\
Ours-FT     & 0.395  & 0.471  & 15.29  & 121.9   & 55.87  & 910.8 \\
Ours-FT + Joint Training   & 0.391  & 0.481  & 16.12  & 120.2  & 53.42  & 891.3 \\
Ours-FT + Full FT    &  0.383  & 0.489  & 16.41  & 118.3  & 53.12  & 924.1 \\
\midrule
Ours          & \textbf{0.381}   &  \textbf{0.497}  &  \textbf{16.80}  &  \textbf{116.6}   &  \textbf{51.69}  &  \textbf{854.9} \\
\bottomrule
\end{tabular}
}
\vspace{-5mm}
\label{tab:main_quantitative}
\end{table}


\noindent\textbf{Ablation Study.}
We test the following variants:
(1) \textit{Ours-FG-MRA-FT}: naive implementation.  
(2) \textit{Naive+RefNet}: naive implementation with ReferenceNet.  
(3) \textit{Ours-MRA-FT}: our model without multi-reference attention and fine-tuning, same as \textit{Naive+FG} in Fig.~\ref{fig:pipeline_ablation}.  
(4) \textit{Ours-FG}: our model without frame generation strategy, where multi-reference attention is modified to use only the first frame rather than the first two.
(5) \textit{Ours-MRA}: our model without multi-reference attention.  
(6) \textit{Ours-RIA}: our model without reference image augmentation.  
(7) \textit{Ours-FT}: our model without fine-tuning, same as \textit{Naive+FG+MRA} in Fig.~\ref{fig:pipeline_ablation}.
\revision{
(8) \textit{Ours-FT+Joint Training}: our model replacing image-only fine-tuning in Sec.~\ref{sec:fine-tuning} with joint fine-tuning on image and video data.
We use the same image data as in Sec.~\ref{sec:fine-tuning} together with our fit check training videos.
(9) \textit{Ours-FT+Full FT}: our model replacing spatial-layer-only fine-tuning in Sec.~\ref{sec:fine-tuning} with full fine-tuning of the denoising network.
}

Tab.~\ref{tab:main_quantitative} and Fig.~\ref{fig:pipeline_ablation} show that our final model outperforms all variants.
Beyond the discussion in Sec.~\ref{sec:method}, we observe: 
(1) \textit{Ours} surpasses \textit{Ours-RIA}, validating the benefit of reference image augmentation.
\revision{
(2) \textit{Ours} outperforms \textit{Ours-FT + Joint Training} and \textit{Ours-FT + Full FT}, demonstrating the advanatge of fine-tuning only the spatial layers.
}
Please see supplementary for additional results.


 \begin{figure}[!t]
\centering
  \subcaptionbox{Input selfies and pose}%
{\includegraphics[width=0.23\linewidth]{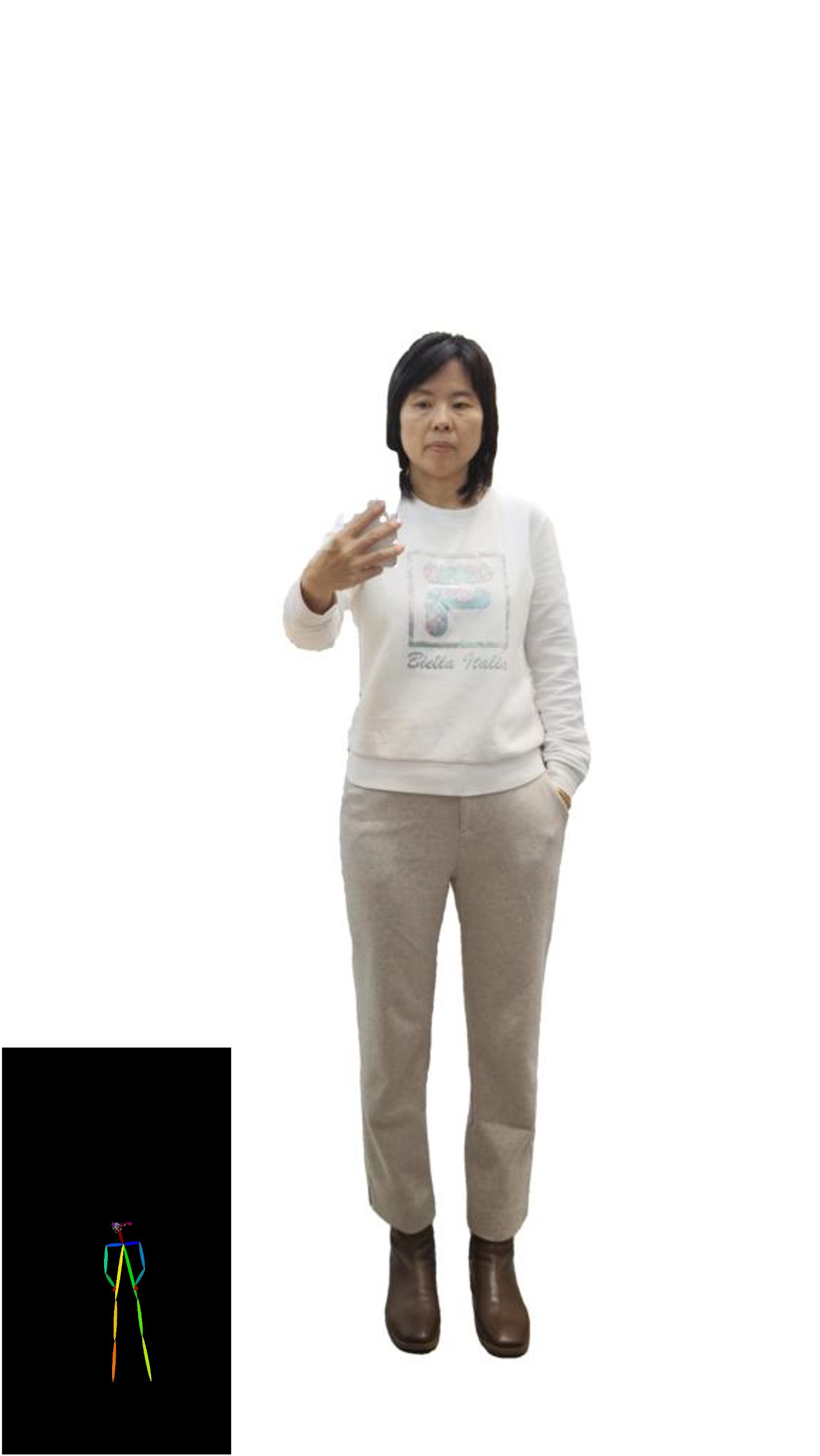}\hfill\includegraphics[width=0.23\linewidth]{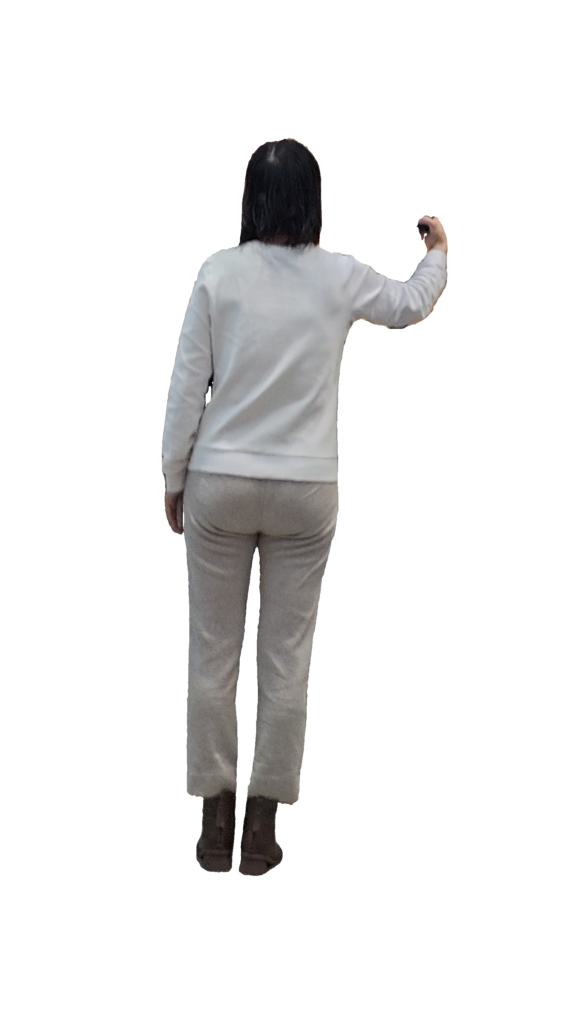}}
  \subcaptionbox{Background}%
{\includegraphics[width=0.23\linewidth]{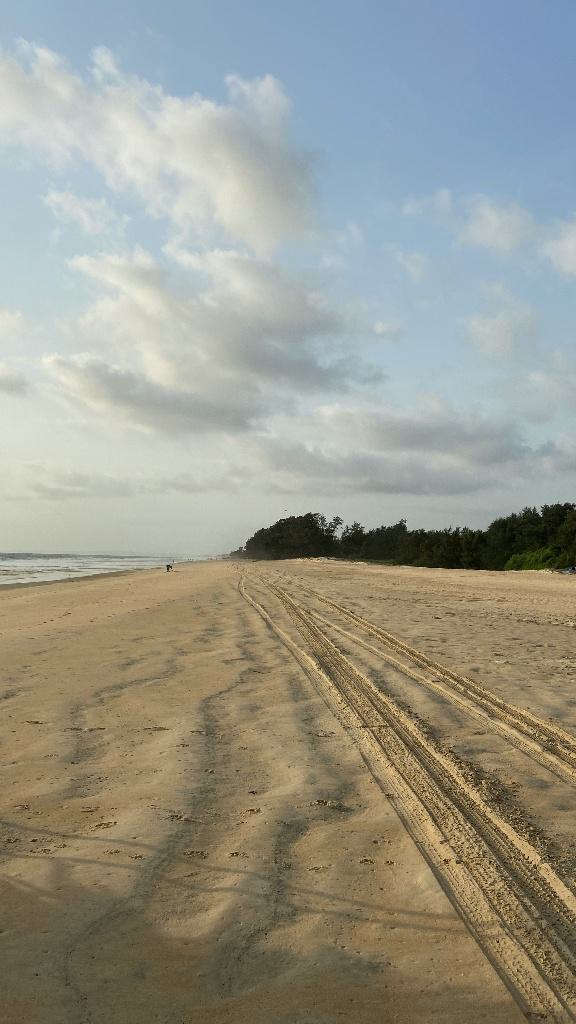}}
  \subcaptionbox{Ours}%
{\includegraphics[width=0.23\linewidth]{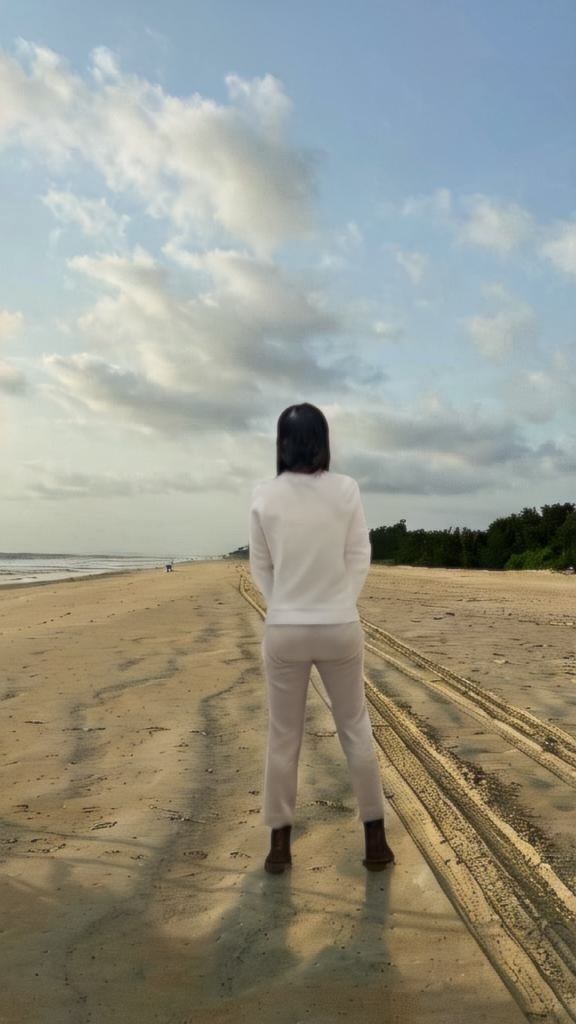}}
%
\vspace{-3mm}
\caption{\textbf{Limitations.} (a) and (b) show the inputs; in (b), the sun direction is not from the front. Our method (c) misestimates the sun direction, causing incorrect shadow.
}
\vspace{-4mm}
  \label{fig:limitations}
\end{figure}

\noindent\textbf{Limitations.}
Our method has several limitations.
First, it fails to generate realistic shadows when compositing subjects into sunlit backgrounds (Fig.~\ref{fig:limitations}).
Second, constrained by the base model of MimicMotion, it cannot faithfully render fine details such as hands and intricate textures.
Third, the background in generated frames might have slight color shift compared to the input background due to the use of VAE and the color shift between frames in  training videos. 
These issues could be mitigated by adopting a more advanced video diffusion model.
Moreover, our motion capture method does not model arm movements due to the single-phone IMU setup; integrating additional IMUs (\eg, from a smartwatch) could address this.
\revision{
Lastly, while the model performs well with the current dataset, scaling up training data can further improve the model.
}

\noindent \textbf{Acknowledgement}.
This work was supported by the UW Reality Lab and Google. 




{
    \small
    \bibliographystyle{ieeenat_fullname}
    \bibliography{main}
}


\end{document}


\maketitle

\section{Motion and Background Retrieval}
\noindent\textbf{Motion Acquisition}.
For motion retrieval, the key idea is to retrieve the top-k closest matches by computing dynamic time warping (DTW) distances between the recorded and candidate motions' orientation and translation.  For orientation, we focus on the horizontal direction (yaw) since a mobile phone's IMU provides a more robust estimate in this axis compared to roll and pitch. Moreover, yaw captures spinning, which, along with translation, effectively represents most of the motions. 
For simplicity, we define orientation as yaw throughout the rest of the paper.
Below we introduce the details of motion retrieval. 

First, the user performs a motion while holding a phone, recording IMU data to obtain orientation $O_{1:\Tilde{N}}$ and global translation $\mathcal{T}_{1:\Tilde{N}}$, where $\Tilde{N}$ is the motion length.  
Here, the orientation is directly derived from the device’s motion sensors, while the global translation is computed through a multi-step process. First, we filter the acceleration data to reduce noise. Then, we integrate the filtered acceleration to obtain velocity, followed by a second integration to derive translation over time.

For each candidate motion $i$ in the database (stored as RGB video), we extract the SMPL~\cite{SMPL-X:2019} sequence $S^i_{1:\bar{N}}$ and global translation $\mathcal{T}^i_{1:\bar{N}}$ using a pretrained human pose estimator~\cite{li2022cliff}, where $\bar{N}$ is the motion length. The candidate's orientation $O^i_{1:\bar{N}}$ is obtained from the yaw component of the global rotation of the root joint in $S^i_{1:\bar{N}}$.
We compute the distance between the recorded motion and each candidate as:
\begin{equation}
    D^i = D_{dtw} (O_{1:\Tilde{N}}, O^i_{1:\bar{N}}) + \alpha D_{dtw} (\mathcal{T}_{1:\Tilde{N}}, \mathcal{T}^i_{1:\bar{N}}),
\end{equation}
where $D_{dtw}$ is dynamic time warping (DTW) to handle sequences of different lengths. $\alpha=0.1$ is a constant that balances orientation and translation importance. All candidates are ranked based on $D^i$, and the top-k motions are retrieved for user selection. After selection, we extract the DWPose sequence $P_{1:N}$ as target pose sequence. 

In practice, we set our motion database to the same as our training set, as the training videos consist of fit check videos, which provide well-suited motions.

\begin{figure}[!t]
\centering
{\includegraphics[width=0.23\linewidth]{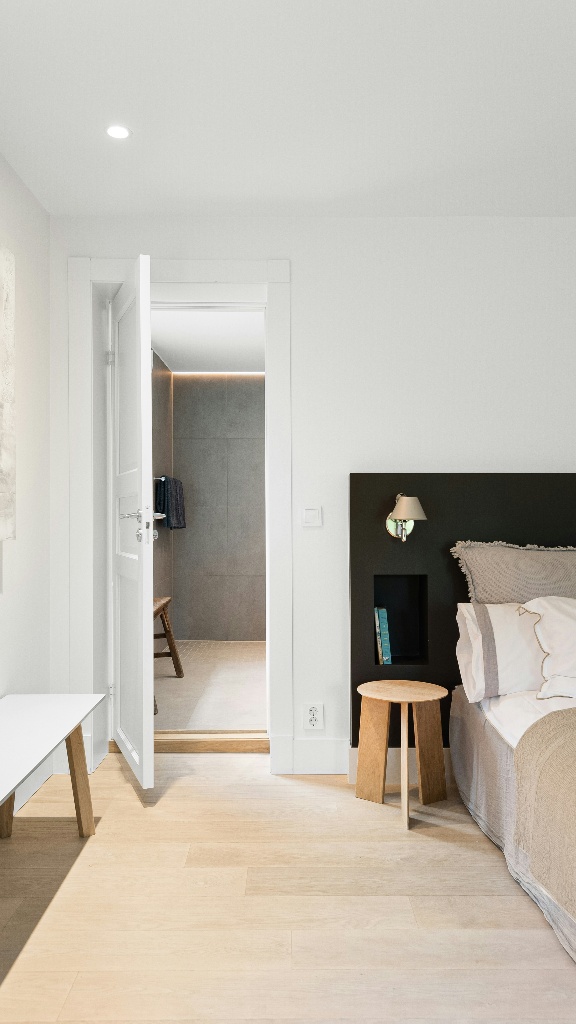}}
{\includegraphics[width=0.23\linewidth]{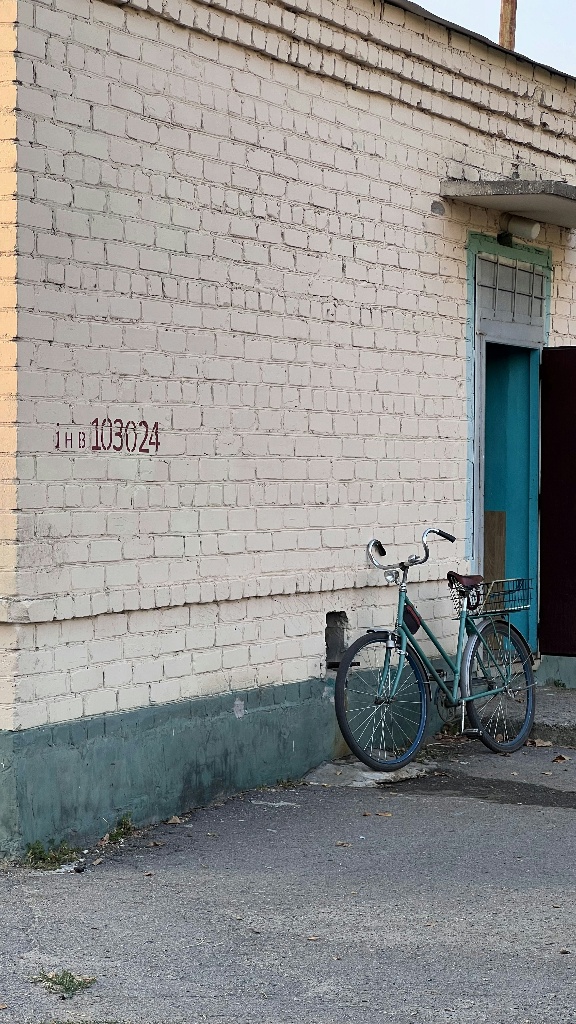}}
{\includegraphics[width=0.23\linewidth]{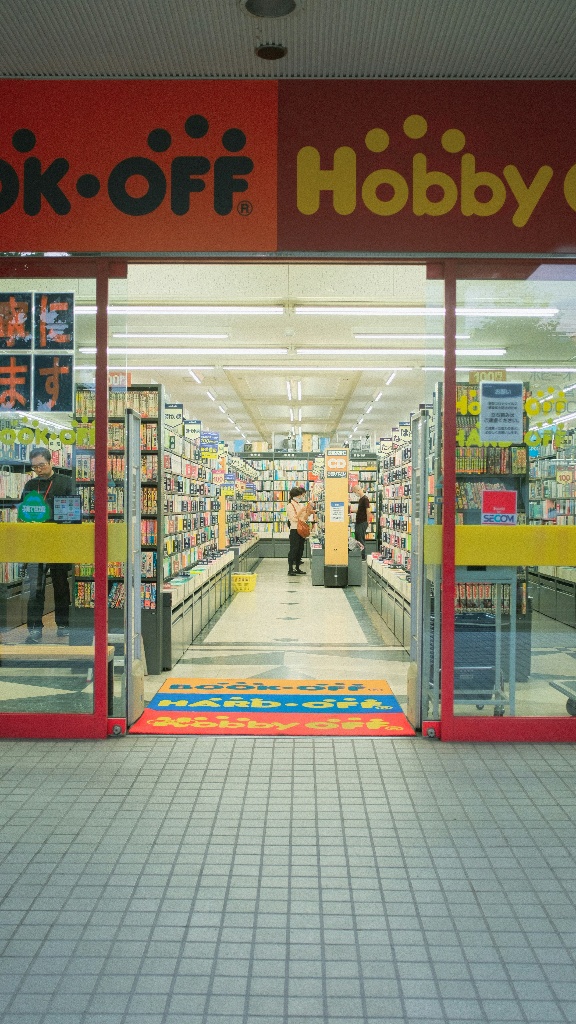}}
{\includegraphics[width=0.23\linewidth]{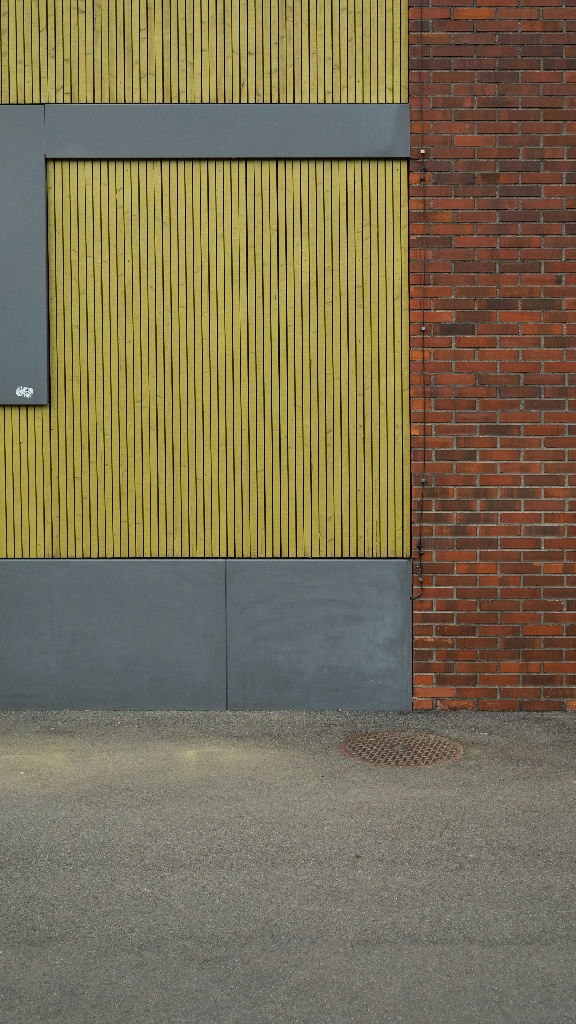}}
\caption{\textbf{Examples of Images from Background Database.} The dataset contains a diverse collection of images spanning both indoor and outdoor environments.
}
  \label{fig:bg_example}
\end{figure}


\noindent\textbf{Background Acquisition}.
The retrieved motion must be compatible with the new background, ensuring natural alignment with the ground plane. One approach is to rotate the SMPL sequence of the retrieved motion to align with the ground plane of any given background, and then extract DWPose from the rendered motion. However, this fails due to foot sliding in the estimated SMPL sequence. Instead, we opt for background retrieval, selecting backgrounds where the ground plane is closely aligned with the retrieved motion. Specifically, we begin by utilizing a pretrained depth estimation model~\cite{wang2024moge} and a pretrained image segmentation model~\cite{jain2022oneformer} to estimate the ground plane normal for both the original background -- where the motion was captured -- and all candidate backgrounds. Then, we retrieve the top-k backgrounds from our database with the closest ground plane normals. Once a background is selected, we automatically scale and translate $P_{1:N}$ to keep foot keypoints on the ground.

For our background database, we curate a diverse database of indoor and outdoor environments, consisting of 800 images, including both AI-generated and real images. Fig.~\ref{fig:bg_example} shows some examples of images from the background database.

\section{Implementation Details}
We will introduce the implementation details below, and all the images and videos are operated in the resolution of height 1024 and width 576, same as the MimicMotion. 
\textit{We will release the code upon acceptance. }

\subsection{Our Method}
\textbf{Reference Image Augmentation.}
During training, we apply a pretrained image harmonization network~\cite{10285123} to adjust the color tone of the front and back reference images, $I'_{fr}$ and $I'_{bk}$. This network takes a composite (unharmonized) image and a foreground mask as input, then produces a harmonized image where the foreground seamlessly blends with the background.

For each reference image, we first apply a pretrained image matting method~\cite{ravi2024sam2} to obtain the foreground mask. We then randomly select a background image from our background database and composite it with the extracted human foreground to create a composite image. The composite image and its corresponding foreground mask are fed into the image harmonization network, which adjusts the foreground color to better match the background. 
After obtaining the harmonized output images, we remove their backgrounds and use the resulting images as training data. We apply this process independently to both $I'_{fr}$ and $I'_{bk}$, using different background images for each. This is to accommodate the natural color tone variations between front and back selfies at test time, even when captured almost simultaneously.

\noindent\textbf{Model Architecture and Parameters.} 
We adopt the 3D denoising UNet, image encoder, pose encoder, VAE encoder, and VAE decoder architectures from MimicMotion~\cite{zhang2024mimicmotion}. Due to computational constraints, we set $T$ to 6, which means each training batch contains 8 frames (including front and back reference image).   Larger $T$ can be used if more computational resource is available.  While we train on 8 video frames per batch, we find that the model can be extended to generate 16-frame or 24-frame sequences at test time, improving efficiency without a noticeable loss in quality. 

\noindent\textbf{Model Training.} 
We initialize the model using the pretrained checkpoint ``MimicMotion\_1.pth'' from MimicMotion.  We did not apply  regional loss amplification in MimicMotion because we found that this does not improve the results in our experiments. 
During training, we randomly sample the front and back view images, $I'_{fr}$ and $I'_{bk}$, from the training video $V'_{1:N}$. 
To sample a $T$-length sequence $V'_{1:T}$ from the training video, we apply the following strategy:
\begin{itemize}
    \item Randomly select frames from the video (20\% of the time)
    \item Select a sequence containing at least one front-facing frame (not necessarily $I'_{fr}$) (40\% of the time)
    \item Select a sequence containing at least one back-facing frame (not necessarily $I'_{bk}$) (40\% of the time)
\end{itemize}
   
Additionally, we apply reference image augmentation to both $I'_{fr}$ and $I'_{bk}$ for 50\% of time.

During training, each conditioning feature -- $f_v$, $f_{im}$ and $f_p$ -- is randomly dropped (set to zero) 10\% of the time, following the classifier-free guidance method~\cite{ho2022classifier}. This allows us to control the strength of each conditional signal during inference.  
Training runs on a single NVIDIA A100 GPU with a batch size of 1 and a learning rate of 1e-5, for 220K steps (around 112 hours).

\begin{figure}[!t]
\centering
{\includegraphics[width=0.23\linewidth]{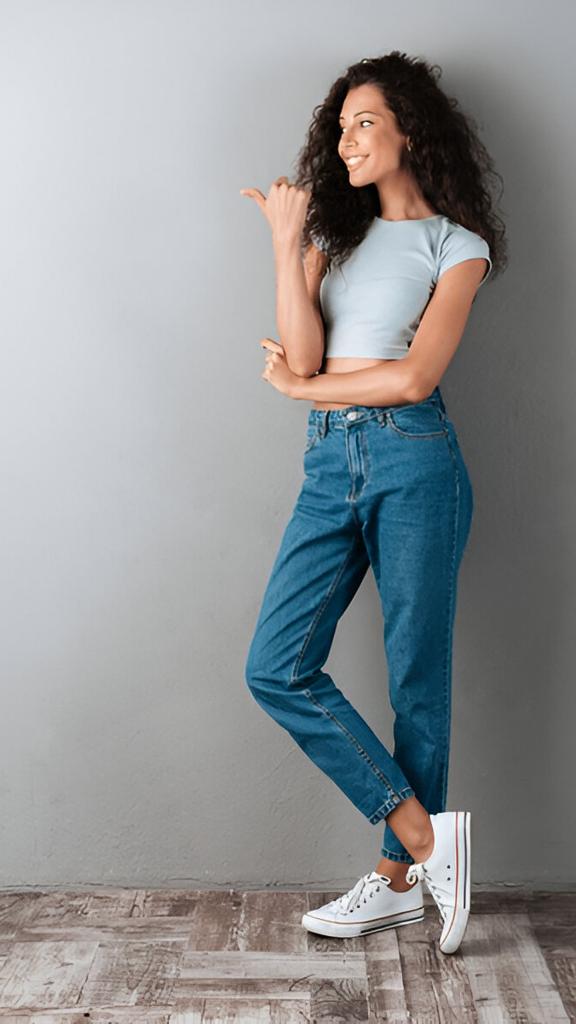}}
{\includegraphics[width=0.23\linewidth]{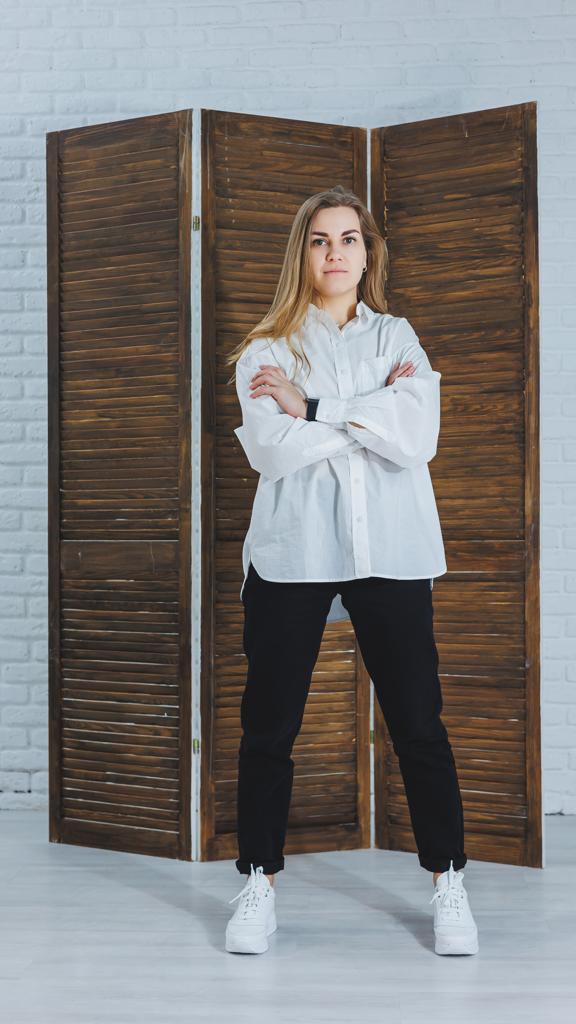}}
{\includegraphics[width=0.23\linewidth]{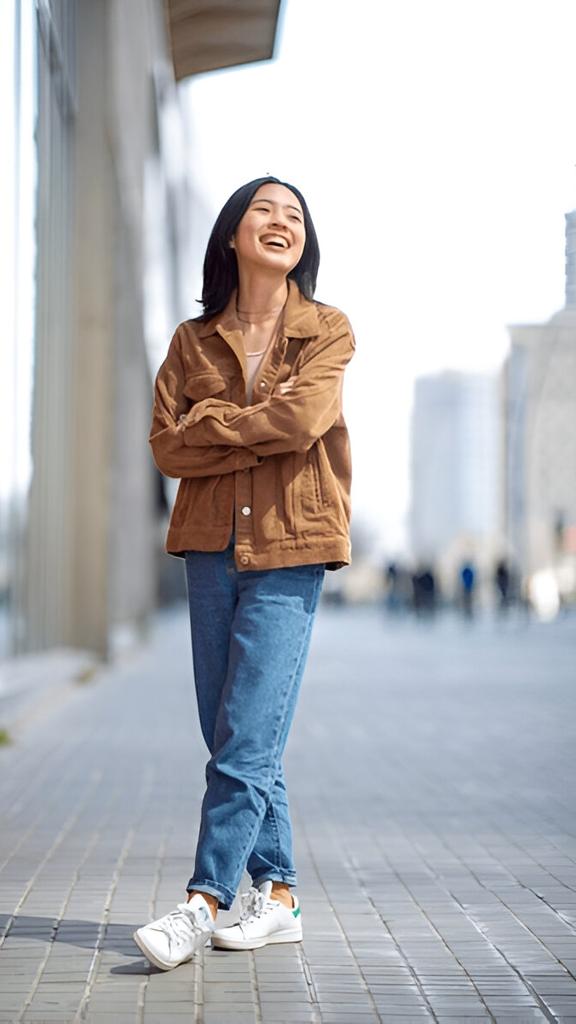}}
{\includegraphics[width=0.23\linewidth]{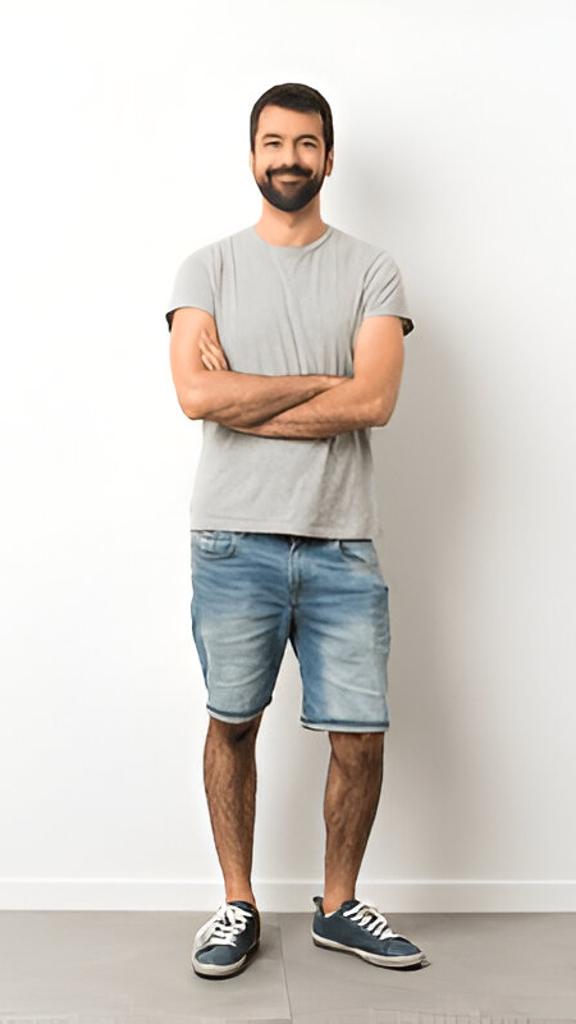}}
\caption{\textbf{Collected Front-Facing Samples for Fine-Tuning.} We collect front-facing images from the web that contain visible shadows or reflections. These images are used to generate data pairs for the fine-tuning dataset.
Fine-tuning on high-quality images enhances sharpness and improves the generation of shadows and reflections.
}
  \label{fig:ft_example}
\end{figure}

\noindent\textbf{Model Fine-Tuning.} 
We fine-tune the trained model on a high-quality image dataset. Fig.~\ref{fig:ft_example} presents examples of the front-facing human images we collected, which are later used to generate data pairs for fine-tuning. The shadow and reflection regions are manually annotated.
During fine-tuning, we omit reference image augmentation, as we observe that the model becomes confused by color tone shifts, resulting in frames with unnatural colors.
We use the same strategy as the training time  to drop the conditioning feature. 
We apply a weighted loss strategy during fine-tuning. Specifically, we assign a higher weight $\beta=2$ to the loss computed in the shadow and reflection regions, while maintaining a weight of 1 for the rest of the region.
For optimization, we use a learning rate of 1e-6 and fine-tune for 1K steps, which takes approximately 30 minutes.

\noindent\textbf{Model Inference.} 
At inference, we set the guidance scale to 2 and the number of overlapping frames to 4.  The denoising time step is set to 25, and we use the Euler scheduler~\cite{karras2022elucidating}.

\subsection{Baseline Details}
\noindent\textbf{Human Animation Baselines.}
We initialize all baselines from their pretrained checkpoints and train them on our dataset on a single NVIDIA A100 GPU for a fair comparison. Additionally, we set the frame length for all baselines to 8, aligning with our settings.

For Animate Anyone\cite{hu2023animateanyone}, since the official code is unavailable, we choose to use a widely-adopted unofficial implementation\cite{moore_animate_anyone}. We follow the same hyperparameter settings as this codebase. The first stage of training runs for 100K steps, taking approximately 40 hours to converge. The second stage runs for 40K steps, requiring around 24 hours to complete. We observe that further training degrades performance.

For Champ~\cite{zhu2024champ}, we use the official code. The first stage of training is conducted for 100K steps, taking approximately 35 hours to converge. The second stage runs for 40K steps, requiring about 20 hours to complete. Similar to Animate Anyone, we find that additional training negatively impacts performance.

For StableAnimator~\cite{tu2024stableanimator}, we follow the official implementation and train the model for 220K steps with a learning rate of 1e-5, taking approximately 132 hours to complete.

For MimicMotion~\cite{zhang2024mimicmotion}, since no training code is provided, we implement our own training procedure. We train the model for 220K steps with a learning rate of 1e-5, which takes around 99 hours to complete.

\noindent\textbf{Motion Retrieval  Baseline.}
As there are no existing IMU-to-motion retrieval baselines, we adopt the text-to-motion retrieval method TMR\cite{petrovich23tmr} as our baseline. We use the official implementation and run the pretrained model (trained on the HumanML3D dataset\cite{Guo_2022_CVPR}) on the recorded motion and our motion database to retrieve the top-k matching motions.

\section{Experiments}

\subsection{More Results for Selfie Input}

\begin{figure*}[!t]
    \centering
\includegraphics[width=0.98\linewidth]{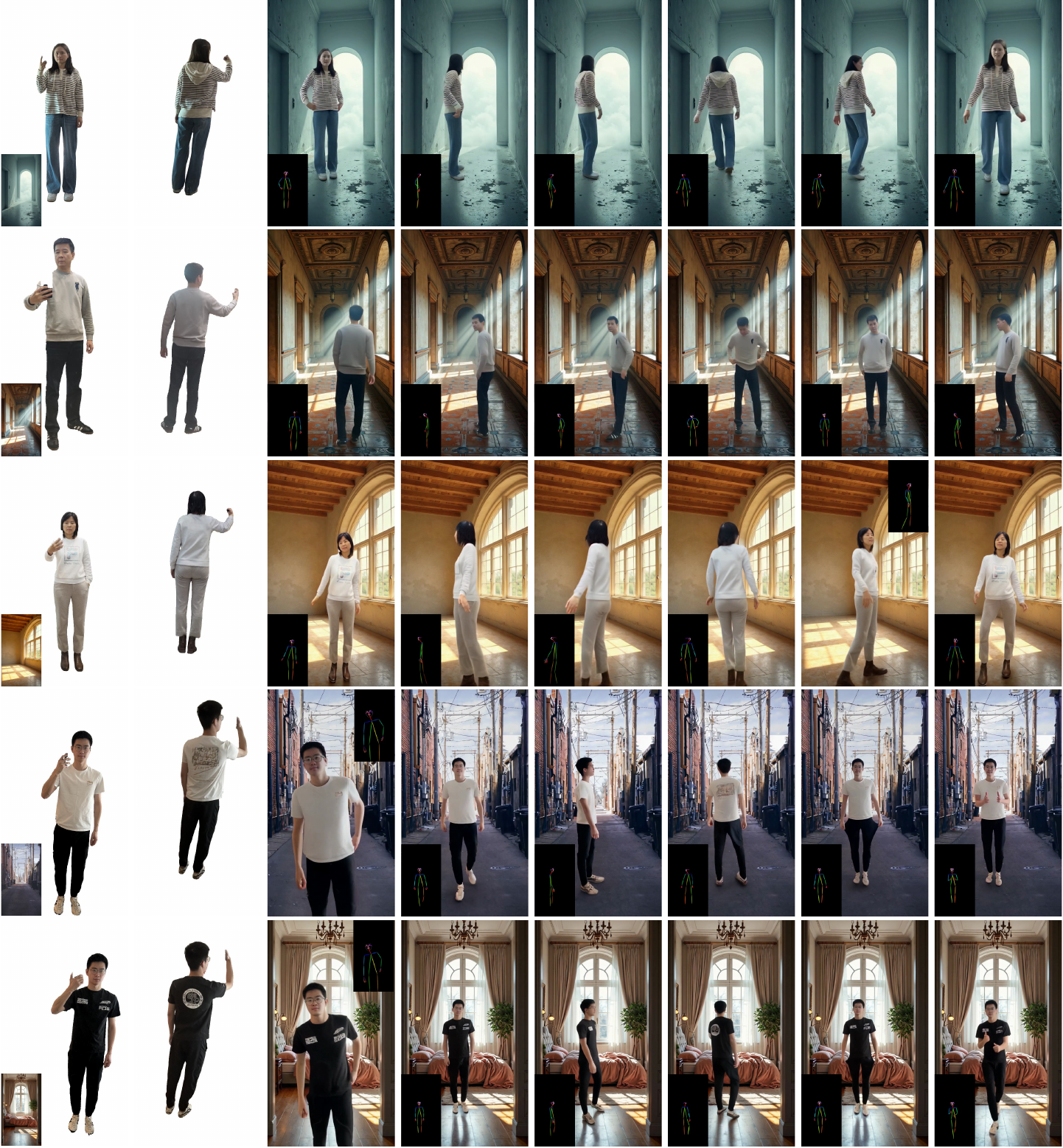}
    \caption{\textbf{More Results of Our Method.} The left two columns display the input selfies and background, while the right six columns show the generated results (inset: pose input, adjusted to prevent occlusion). Given mirror selfies with various outfits and lighting conditions, our method produces realistic fit-check videos, accurately capturing appearance from diverse poses. It also generates reflections and shadows on the ground, ensuring seamless integration between the subject and both indoor and outdoor backgrounds.
    }
    \label{fig:main_results_supp1}
\end{figure*}

Fig. \ref{fig:main_results_supp1} presents additional results of our method on real selfie captures. Our approach generates high-quality fit check videos featuring diverse outfits in both indoor and outdoor settings, accurately capturing a wide range of poses with realistic shading, reflections, and shadows.

\subsection{Comparison with Baseline}

\subsubsection{Datasets}

In addition to evaluating our model on the self-captured real selfie dataset presented in the main paper, we conduct further analysis on three additional test sets: (1) the test set from our self-collected dataset, (2) the test set  of UBC Fashion dataset~\cite{zablotskaia2019dubc_dataset}, and (3) the TikTok dataset~\cite{Jafarian_2021_CVPR_TikTok_dataset}. 
We filter out videos that do not include a back view. After filtering, our dataset test set contains 149 videos, with an average of 68 frames per video. The UBC Fashion dataset consists of 100 videos, averaging 98 frames per video, while the TikTok dataset includes 19 videos, with an average of 115 frames per video.

For evaluation, we randomly sample a front and back image as reference inputs, while the input pose sequence is extracted from the corresponding ground truth (GT) video.
The input backgrounds in our test set and the TikTok dataset are obtained using the same inpainting strategy as in our training set. However, for the UBC Fashion dataset, we use a plain white background instead, as inpainting a nearly white background with Stable Diffusion introduces artifacts.
Finally, we compare the generated video with the GT video.

Notably, the input reference images in these datasets are captured from a third-person perspective rather than as selfies. This setup enables us to evaluate model performance on non-selfie inputs. Furthermore, since these images are sampled from the GT video, they share similar lighting conditions with the GT. This also differs from our selfie setup, but we still evaluate on these datasets for a more comprehensive evaluation.  

Please note that all methods are trained solely on our training set, with baselines initialized from their official checkpoints.

\begin{figure}[!t]
    \centering
\includegraphics[scale=0.34]{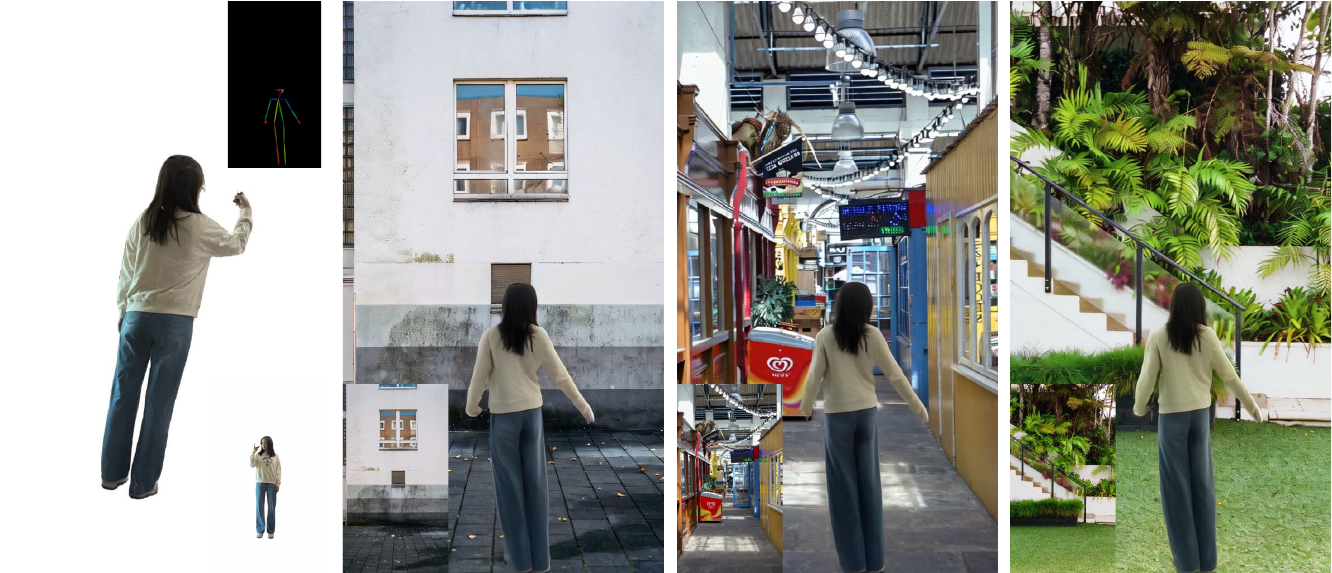}  
    \caption{\textbf{Results with the Same Capture under Different Virtual Backgrounds.} The first column shows the input selfies and target pose; the others show results under different virtual backgrounds (insets: input backgrounds). Despite strong left lighting in the selfies, our method adapts shading to each background.
    }
    \label{fig:diff_bg_supp}
\end{figure}

\subsubsection{Qualitative Comparison}

Fig.~\ref{fig:diff_bg_supp} shows more results of the same captures under different virtual backgrounds.

\begin{figure*}[!t]
    \centering
\includegraphics[width=0.99\linewidth]{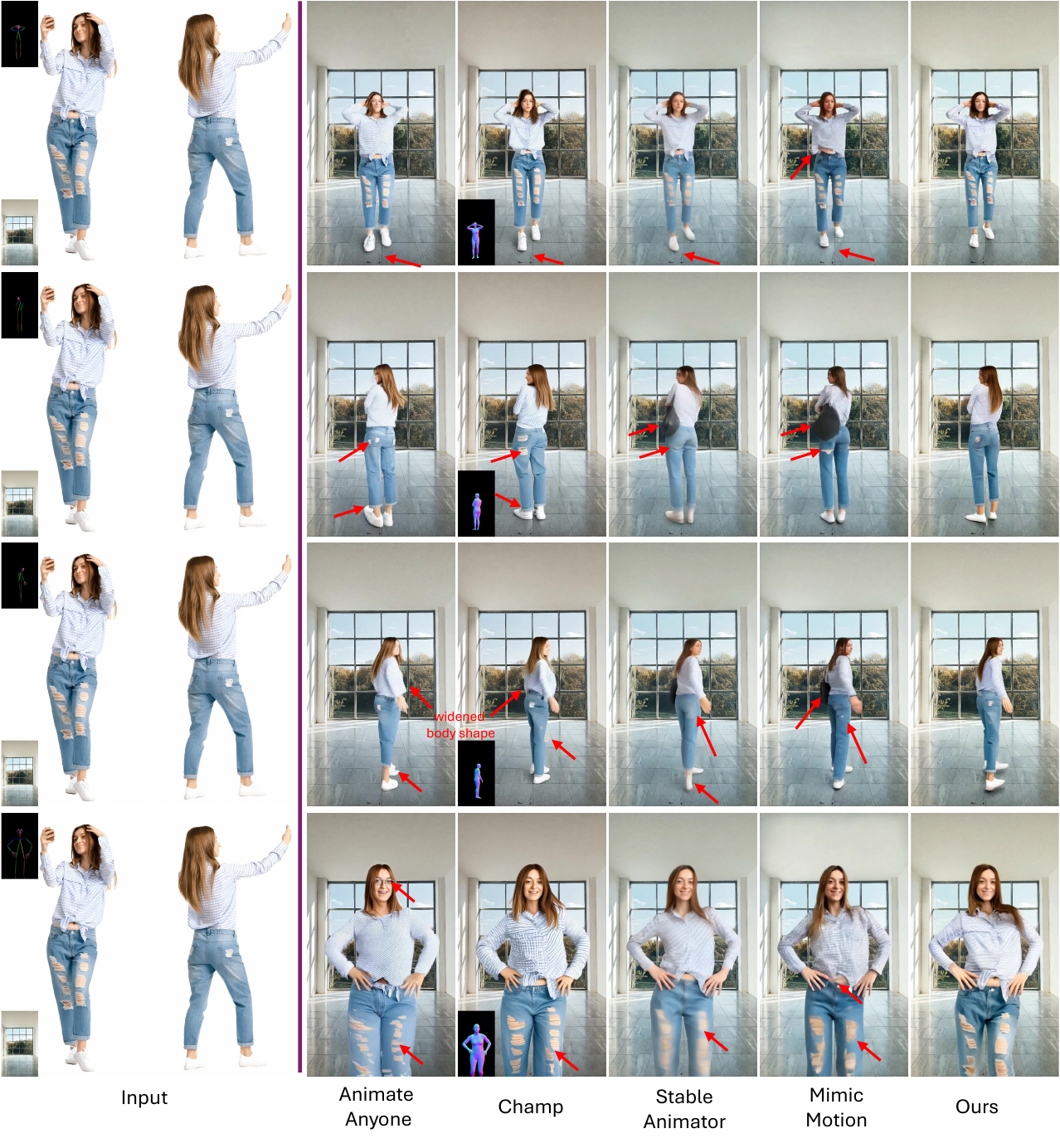}
    \caption{\textbf{Qualitative Comparison with Baselines on Selfie Inputs.} The left two columns display the input selfies, pose, and background, while the right five columns showcase the generated results from various methods. The inset of Champ illustrates its corresponding SMPL pose input. All results are post-processed using face refinement. Our method surpasses all tested baselines by more accurately reconstructing appearance across diverse poses while also producing more realistic shadows and reflections on the floor. 
    }
    \label{fig:comparison_self1}
\end{figure*}

\begin{figure*}[!t]
    \centering
\includegraphics[width=0.99\linewidth]{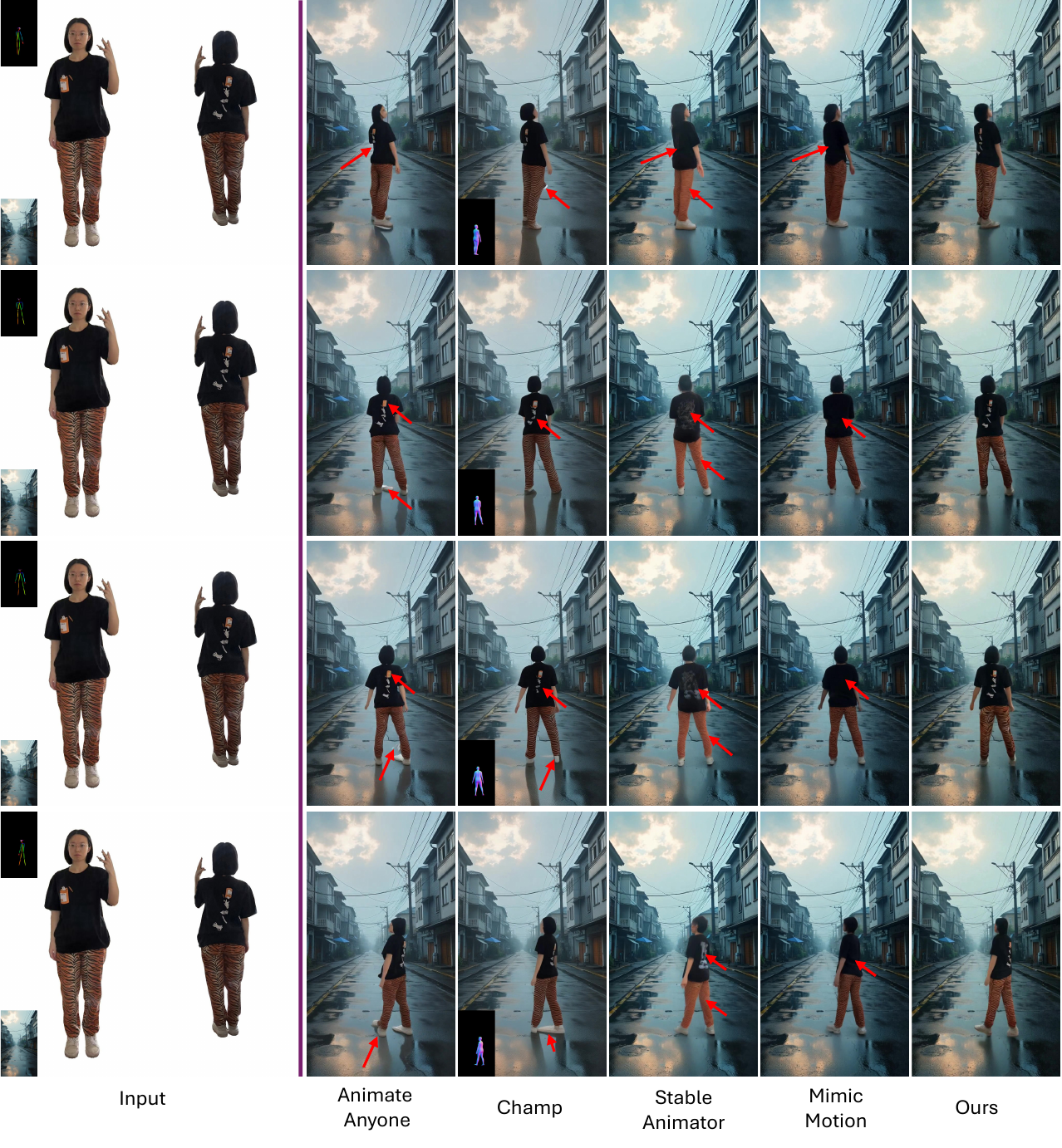}
    \caption{\textbf{Qualitative Comparison with Baselines on Selfie Inputs.} The left two columns display the input selfies, pose, and background, while the right five columns showcase the generated results from various methods. The inset of Champ illustrates its corresponding SMPL pose input. All results are post-processed using face refinement. Our method surpasses all tested baselines by more accurately reconstructing appearance across diverse poses while also producing more realistic shadows and reflections on the floor. 
    }
    \label{fig:comparison_self2}
\end{figure*}

\begin{figure*}[!t]
    \centering
\includegraphics[width=0.95\linewidth]{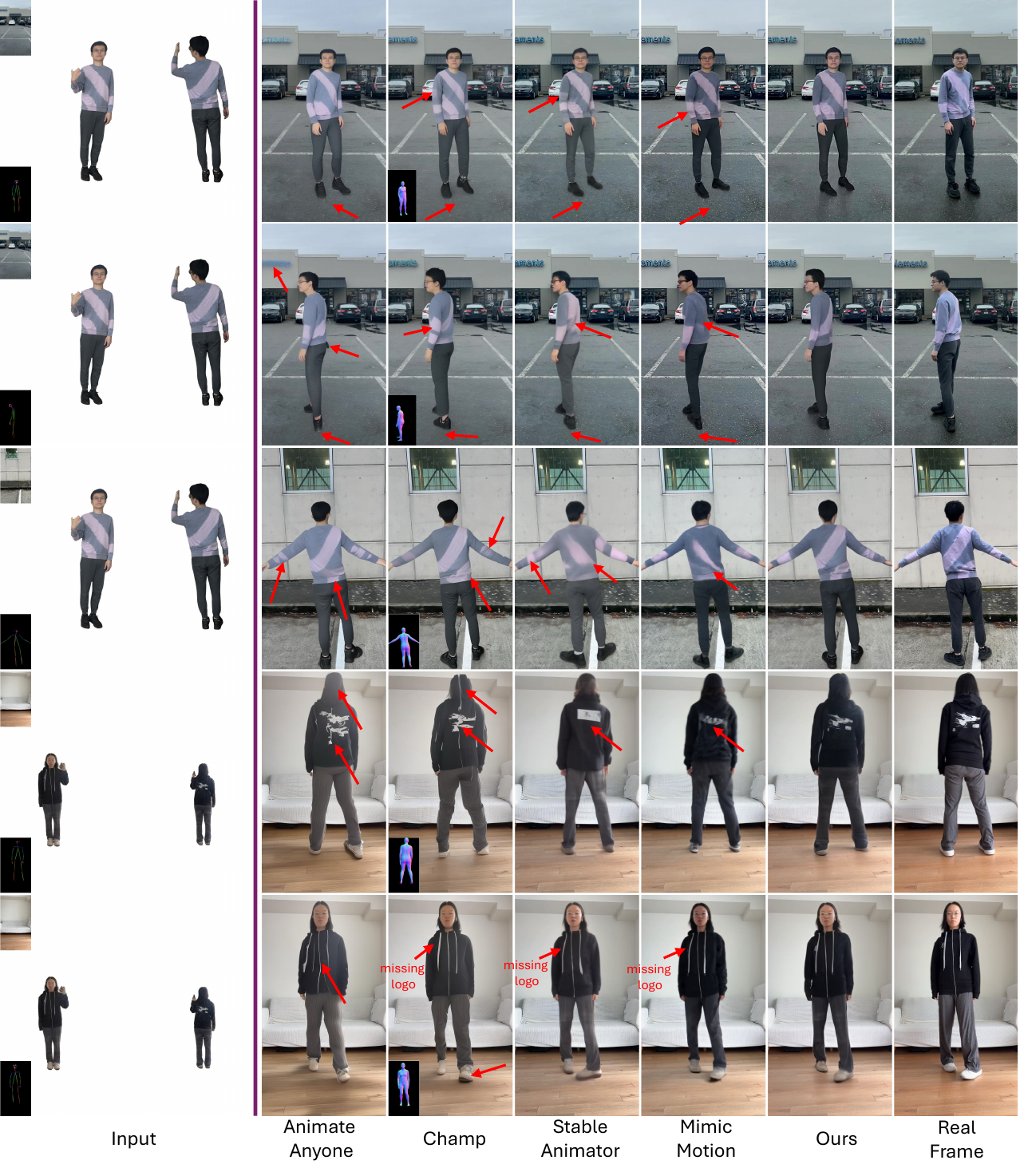}
    \caption{\textbf{Qualitative Comparison with Baselines on Self-Captured Dataset.} The left two columns display the input selfies, pose, and background, while the right six columns showcase the generated results from various methods alongside the real frame. The target poses are detected from the real frames. The inset of Champ illustrates its corresponding SMPL pose input. All results are post-processed using face refinement. Our method surpasses all tested baselines by more accurately reconstructing appearance across diverse poses while also producing more realistic shadows and reflections on the floor. Note that background images and real videos, though captured in the same session, may vary in intensity and color tone due to auto-exposure and white balance adjustments.
    }
    \label{fig:comparison_self_gt}
\end{figure*}

Fig.~\ref{fig:comparison_self1}, Fig.~\ref{fig:comparison_self2}, and Fig.~\ref{fig:comparison_self_gt} present additional comparisons on real selfie captures. We observe the following:
(1) Animate Anyone and Champ exhibit artifacts such as inaccurate clothing patterns and artifacts around the shoes (row 3, column 3 in Fig.~\ref{fig:comparison_self2}).
(2) MimicMotion and StableAnimator fail to accurately reconstruct the appearance of the back view, demonstrating that simple modifications to these methods do not effectively utilize the additional reference image input. Additionally, they struggle to capture fine details in the front view (\eg missing logo in row 5 in Fig.~\ref{fig:comparison_self_gt}) and produce blurry patterns, whereas our method preserves accurate and sharp patterns due to the fine-tuning stage.
(3) Our method surpasses all baselines in both appearance and pose fidelity while also generating more realistic reflections and shadows on the floor.

\begin{figure*}[!t]
    \centering
\includegraphics[width=0.95\linewidth]{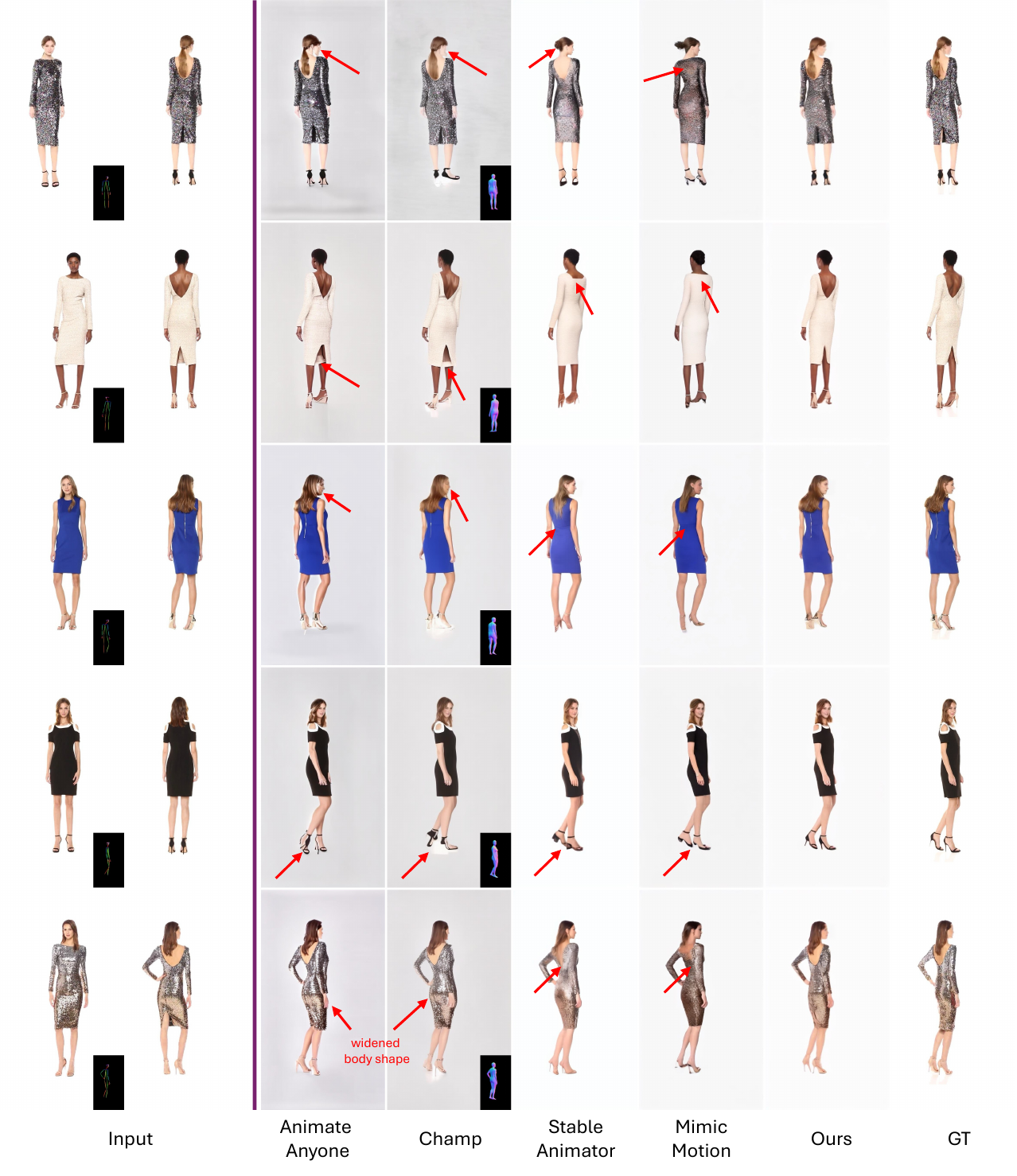}
    \caption{\textbf{Qualitative Comparison with Baselines on UBC Fashion Dataset.} The left two columns present the input reference images, pose, and background, while the right six columns showcase the generated results from various methods alongside the ground truth (GT). The target poses are detected from the GT. The inset of Champ displays its corresponding SMPL pose input. All results are post-processed using face refinement.
 Note that none of the methods were trained on the UBC Fashion Dataset's training set. 
 Our method outperforms all tested baselines in accurately reconstructing appearance across a diverse range of poses.
    }
    \label{fig:comparison_ubc}
\end{figure*}

Fig.~\ref{fig:comparison_ubc} presents a qualitative comparison on the UBC Fashion dataset. We observe the following:
(1) Image diffusion-based methods (Animate Anyone and Champ) exhibit noticeable background color shifts, indicating their limited generalization ability to unseen backgrounds. Additionally, they introduce visible artifacts on faces and bodies and fail to accurately capture body shape.
(2) Video diffusion-based baselines (MimicMotion and StableAnimator) struggle to reconstruct the appearance of the back view accurately, highlighting that simple modifications to these methods do not effectively utilize the additional reference image input.
(3) Our method outperforms all baselines in both appearance and pose fidelity, producing more realistic and coherent results.

\begin{figure*}[!t]
    \centering
\includegraphics[width=0.95\linewidth]{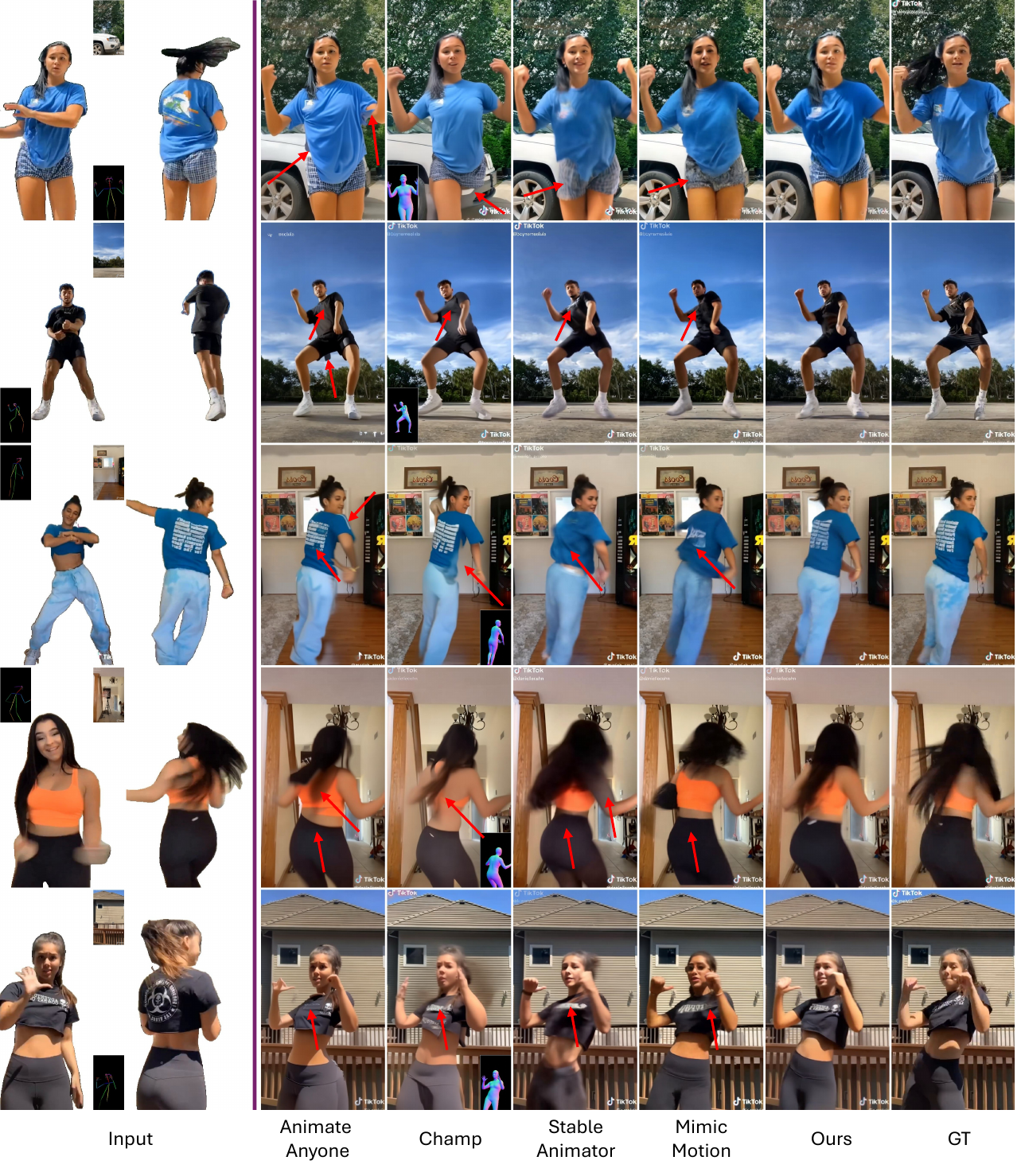}
    \caption{\textbf{Qualitative Comparison with Baselines on TikTok Dataset.} The left two columns present the input reference images, pose, and background, while the right six columns showcase the generated results from various methods alongside the ground truth (GT). The target poses are detected from the GT. The inset of Champ displays its corresponding SMPL pose input. Note that none of the methods were trained on the TikTok Dataset. All results are post-processed using face refinement.  Our method outperforms all tested baselines in accurately reconstructing appearance across a diverse range of poses.  \textit{Disclaimer:  In Fig. 2 of the main paper, the example in row 3 of this figure is used solely for illustrative purposes and was not included in the training data.  }
    }
    \label{fig:comparison_tiktok}
\end{figure*}

Fig.~\ref{fig:comparison_tiktok} presents a qualitative comparison on the TikTok dataset. We observe the following:
(1) Animate Anyone and Champ exhibit noticeable artifacts, such as missing body parts (e.g., row 1, column 4, and row 3, column 4) and inaccurate clothing patterns (rows 2 to 5).
(2) MimicMotion and StableAnimator struggle to accurately reconstruct the appearance of the back view, demonstrating that simple modifications to these methods do not effectively utilize the additional reference image input. Additionally, they produce blurry patterns (e.g., shorts in row 1), whereas our method generates sharper details due to the design of the fine-tuning stage.
(3) Our method surpasses all baselines in both appearance and pose fidelity, delivering more realistic and coherent results.

\begin{table}[!t]
\centering
\caption{\textbf{Quantitative Comparisons on Our Test Set.} All methods are evaluated without face refinement as post-processing. For each metric, the best and second-best methods are highlighted in bold and underline, respectively.
Our method outperforms all tested baselines across all metrics. The ablation variant, \textit{Ours-RIA}, achieves results comparable to \textit{Ours} on this dataset because the input front and back images share the same background as the ground truth (GT) frames, making reference image augmentation less necessary in this case.
}
\resizebox{0.47\textwidth}{!}{
\begin{tabular}{|l|cccccc|}
\toprule
{Method} & {SSIM} $\uparrow$ & {LPIPS} $\downarrow$ & {PSNR} $\uparrow$ & {FID} $\downarrow$ & {FVD-VID} $\downarrow$ & {FVD} $\downarrow$ \\
\midrule
Animate Anyone~\cite{hu2023animateanyone}  & 0.733 & 0.242 & 19.60 & 128.8 & 57.43 & 421.7 \\
Champ~\cite{zhu2024champ} & 0.763 & 0.231 & 20.60 & 91.72 & 32.40 & 372.9 \\
StableAnimator~\cite{tu2024stableanimator} & 0.771 & 0.219 & 21.03 & 97.26 & 33.33 & 394.7 \\
MimicMotion~\cite{zhang2024mimicmotion} & 0.779 & 0.211 & 21.34 & 88.02 & 30.51 & 366.1 \\
\midrule
Ours-FG-MRA-FT & 0.776 & 0.205 & 21.20 & 89.19 & 29.30 & 356.6 \\
Naive+RefNet & 0.781 & 0.202 & 22.38 & 87.52 & 29.40 & 342.4 \\
Ours-MRA-FT & 0.782 & 0.198 & 22.64 & 86.74 & 28.16 & 308.7 \\
Ours-FG & 0.781 & 0.203 & 22.40 & 87.64 & 25.76 & 311.0 \\
Ours-MRA & \underline{0.796} & 0.186 & 23.08 & 82.82 & 23.71 & 292.2 \\
Ours-RIA & 0.795 & \underline{0.184} & \underline{23.15} & \underline{79.07} & \textbf{22.44} & \underline{281.5} \\
Ours-FT & 0.786 & 0.196 & 22.81 & 86.68 & 26.00 & 298.0 \\
Ours-FT + Joint Training   & 0.789  &  0.188 & 23.01  & 83.22  & 26.42  & 289.3 \\
Ours-FT + Full FT    &  0.792  & 0.186  & 23.07  & 81.52  & 25.42   & 324.1 \\
\midrule
Ours & \textbf{0.799} & \textbf{0.183} & \textbf{23.61} & \textbf{79.02} & \underline{23.11} & \textbf{279.3} \\
\bottomrule
\end{tabular}
}
\label{tab:test}
\end{table}

\begin{table}[!t]
\caption{\textbf{Quantitative Comparisons on the UBC Fashion Dataset.} All methods are evaluated without face refinement as post-processing. For each metric, the best and second-best methods are highlighted in bold and underline, respectively.
Our method outperforms all tested baselines across all metrics. The ablation variant, \textit{Ours-RIA}, achieves results comparable to \textit{Ours} on this dataset because the input front and back images share the same background as the ground truth (GT) frames, making reference image augmentation less necessary in this case.
}
\centering
\resizebox{0.47\textwidth}{!}{
\begin{tabular}{|l|cccccc|}
\toprule
{Method} & {SSIM} $\uparrow$ & {LPIPS} $\downarrow$ & {PSNR} $\uparrow$ & {FID} $\downarrow$ & {FVD-VID} $\downarrow$ & {FVD} $\downarrow$ \\
\midrule
Animate Anyone~\cite{hu2023animateanyone}   & 0.902 & 0.130 & 17.08 & 60.27 & 51.94 & 480.0 \\
Champ~\cite{zhu2024champ} & 0.888 & 0.130 & 16.43 & 59.34 & 56.06 & 363.7 \\
StableAnimator~\cite{tu2024stableanimator} & 0.914 & 0.071 & 21.61 & 58.47 & 31.67 & 191.5 \\
MimicMotion~\cite{zhang2024mimicmotion} & 0.918 & 0.069 & 22.05 & 56.27 & 28.04 & 185.7 \\
\midrule
Ours-FG-MRA-FT & 0.922 & 0.065 & 21.76 & 56.55 & 27.88 & 181.7 \\
Naive+RefNet & 0.922 & 0.064 & 22.47 & 49.18 & 24.64 & 183.1 \\
Ours-MRA-FT & 0.924 & 0.061 & 22.58 & 48.99 & 21.02 & 170.7 \\
Ours-FG & 0.921 & 0.068 & 21.94 & 53.17 & 28.13 & 169.4 \\
Ours-MRA & 0.928 & \underline{0.055} & \underline{23.68} & 47.41 & 13.64 & 149.8 \\
Ours-RIA & \underline{0.932} & \underline{0.055} & 23.59 & \underline{46.42} & \underline{12.70} & \underline{144.2} \\
Ours-FT & 0.925 & 0.057 & 23.39 & 48.68 & 19.31 & 166.0 \\
Ours-FT + Joint Training   & 0.923  &  0.059 & 23.54 & 48.32  & 20.14  & 148.7 \\
Ours-FT + Full FT    &  0.928  & 0.057  & 23.66  & 47.39  & 18.95  & 174.2 \\
\midrule
Ours & \textbf{0.937} & \textbf{0.052} & \textbf{23.73} & \textbf{45.33} & \textbf{12.36} & \textbf{138.7} \\
\bottomrule
\end{tabular}
}
\label{tab:ubc}
\end{table}

\begin{table}[!t]
\centering
\caption{\textbf{Quantitative Comparisons on the TikTok Dataset.} All methods are evaluated without face refinement as post-processing. For each metric, the best and second-best methods are highlighted in bold and underline, respectively.
Our method outperforms all tested baselines across all metrics. The ablation variant, \textit{Ours-RIA}, achieves results comparable to \textit{Ours} on this dataset because the input front and back images share the same background as the ground truth (GT) frames, making reference image augmentation less necessary in this case.
}
\resizebox{0.47\textwidth}{!}{
\begin{tabular}{|l|cccccc|}
\toprule
{Method} & {SSIM} $\uparrow$ & {LPIPS} $\downarrow$ & {PSNR} $\uparrow$ & {FID} $\downarrow$ & {FVD-VID} $\downarrow$ & {FVD} $\downarrow$ \\
\midrule
Animate Anyone~\cite{hu2023animateanyone}   & 0.779 & 0.236 & 18.78 & 81.02 & 44.55 & 551.9 \\
Champ~\cite{zhu2024champ} & 0.774 & 0.243 & 18.24 & 90.89 & 53.81 & 669.2 \\
StableAnimator~\cite{tu2024stableanimator} & 0.784 & 0.242 & 18.48 & 90.43 & 46.20 & 473.4 \\
MimicMotion~\cite{zhang2024mimicmotion} & 0.787 & 0.235 & 18.67 & 87.22 & 38.14 & 433.5 \\
\midrule
Ours-FG-MRA-FT & 0.790 & 0.234 & 18.64 & 87.36 & 37.54 & 435.6 \\
Naive+RefNet & 0.795 & 0.226 & 18.72 & 87.11 & 37.15 & 433.9 \\
Ours-MRA-FT & 0.802 & 0.224 & 18.98 & 85.19 & 37.22 & 436.3 \\
Ours-FG & 0.799 & 0.229 & 18.39 & 81.96 & 36.11 & 399.8 \\
Ours-MRA & 0.805 & 0.217 & 19.42 & 78.60 & 32.35 & 387.5 \\
Ours-RIA & \textbf{0.808} & \underline{0.216} & \textbf{19.71} & \underline{76.70} & \textbf{30.85} & \underline{384.5} \\
Ours-FT & 0.803 & 0.221 & 19.33 & 83.58 & 35.77 & 390.8 \\
Ours-FT + Joint Training   & 0.794  &  0.221 & 19.13 & 79.48  & 34.21  & 390.2\\
Ours-FT + Full FT    &  0.801  & 0.219  & 19.51  &  78.79   & 33.85  & 428.5 \\
\midrule
Ours & \underline{0.807} & \textbf{0.215} & \underline{19.65} & \textbf{76.65} & \underline{31.21} & \textbf{382.1} \\
\bottomrule
\end{tabular}
}
\label{tab:tiktok}
\end{table}

\subsubsection{Quantitative Comparison}

Tab.~\ref{tab:test}, \ref{tab:ubc}, and \ref{tab:tiktok} present the quantitative results of our model compared to the baselines across the three datasets. We observe that Champ performs competitively among the baselines on our test set in terms of video-related metrics, FVD-VID and FVD, but performs worse on the other two datasets. This indicates that this image diffusion-based method achieves better temporal consistency when the input reference images and background are in-distribution. However, its performance degrades significantly for out-of-distribution inputs, demonstrating poor generalization ability.

As discussed in the main paper, our method outperforms all baselines on all metrics by employing a novel frame generation strategy with multi-reference attention and a fine-tuning approach, leading to enhanced appearance fidelity and frame quality. 

\subsection{Body Size, Garment Accuracy, Realism}
Tab.~\ref{tab:huamn_study} reports the full results of our human study, which evaluates body size, garment accuracy, and overall realism. Our method consistently outperforms all baselines across all criteria.

Tab.~\ref{tab:quant_shape} presents the corresponding automatic evaluation results for body size, garment accuracy, and realism. Detailed descriptions and discussions are provided below.

For quantitative body size evaluation, we used the SMPL-based estimator CLIFF~\cite{li2022cliff} to estimate shape parameters on predicted and real frames from our self-captured dataset. We calculated shape difference by averaging the absolute differences between shape parameters of predicted and real frames. Our method achieved a shape difference of 0.053, outperforming the best baseline, MimicMotion (0.062), by 17\%.

We evaluated garment accuracy via region alignment and appearance similarity. For region alignment, we used SAM2~\cite{ravi2024sam2} to segment garment regions in predicted and ground-truth frames from self-captured dataset and computed the average IoU. Our method achieved 0.923, better than the best baseline, MimicMotion (0.895). For appearance similarity, we calculated LPIPS within the segmented garment regions, with our method scoring 0.485, outperforming the best baseline, Champ (0.512).

To evaluate realism, we used a VLM~\cite{Qwen2.5-VL} to rate generated videos on a 1–10 scale (higher is better). Our method scored 7.5, outperforming the best baseline, MimicMotion (6.7). 

\begin{table}[!t]
\centering
\caption{\textbf{Results of Human Study.}  Our method outperforms all baselines.
}
\resizebox{0.5\textwidth}{!}{
\begin{tabular}{|l|ccc|}
\toprule
{Method} & Body Shape Accuracy $\uparrow$ & Garment Accuracy  $\uparrow$  & Realism $\uparrow$ \\
\midrule
Animate Anyone~\cite{hu2023animateanyone}  & 2.55 & 2.45 & 2.03 \\
Champ~\cite{zhu2024champ} & 3.02 &  2.88 & 2.32 \\
StableAnimator~\cite{tu2024stableanimator} & 3.10 & 2.18 & 2.76 \\
MimicMotion~\cite{zhang2024mimicmotion} & 3.20 & 2.05 & 2.82  \\
\midrule
Ours & \textbf{3.88} & \textbf{4.08} & \textbf{3.75} \\
\bottomrule
\end{tabular}
}
\label{tab:huamn_study}
\end{table}

\begin{table}[!t]
\centering
\caption{\textbf{Results of Quantitative Evaluation of Body Size, Garment Accuracy, and Realism.}  Our method outperforms all baselines.
}
\resizebox{0.5\textwidth}{!}{
\begin{tabular}{|l|cccc|}
\toprule
Method 
& \makecell{ Shape Parameter \\ Difference $\downarrow$} 
& \makecell{Garment Region \\  IoU $\uparrow$} 
& \makecell{Garment Appearance \\ LPIPS $\downarrow$}  
&  \makecell{VLM Realism \\ Score $\uparrow$}  \\
\midrule
Animate Anyone~\cite{hu2023animateanyone}  & 0.073 & 0.853 & 0.539  & 5.8 \\
Champ~\cite{zhu2024champ} & 0.075 &  0.861 & 0.512  & 5.6 \\
StableAnimator~\cite{tu2024stableanimator} & 0.071 & 0.884 &  0.616  & 6.2 \\
MimicMotion~\cite{zhang2024mimicmotion} & 0.062 & 0.895 & 0.601 & 6.7  \\
\midrule
Ours & \textbf{0.053} & \textbf{0.923} & \textbf{0.485} & \textbf{7.5} \\
\bottomrule
\end{tabular}
}
\label{tab:quant_shape}
\end{table}

\subsection{Comparison to Models Trained on Other Datasets}
We further compare our method with models trained on different datasets: Veo 3.1 (References to Video)~\cite{google_veo3_1_2025} and Phantom (built on Wan2.1~\cite{wan2025})~\cite{liu2025phantom}. These models take one or multiple reference images and a text prompt as input to generate a video. In our experiments, we provide three reference images -- the front selfie, back selfie, and background image -- along with a text description of motion to generate fit-check videos.

We use the following text prompt for Veo 3.1 (References to Video):
\textit{``Generate a fit-check video. The person [specify the motion]. The person’s appearance must remain consistent with the first and second images (front and back selfies), and the background should match the third image. The camera should remain static and the viewpoint must not change.
''}

We additionally include a background description in the text input for Phantom, as it tends to ignore the provided background image. The text prompt is:
\textit{``Generate a fit-check video. The person [specify the motion]. The person’s appearance must remain consistent with the first and second images (front and back selfies), and the background should match the third image. The background is [description of background]. The camera should remain static and the viewpoint must not change.
''}

Note that, unlike our method, which takes a motion sequence as an explicit input, these models rely solely on text to determine motion. Therefore, their generated motions may differ from ours.

\begin{figure*}[!t]
    \centering
\includegraphics[width=0.95\linewidth]{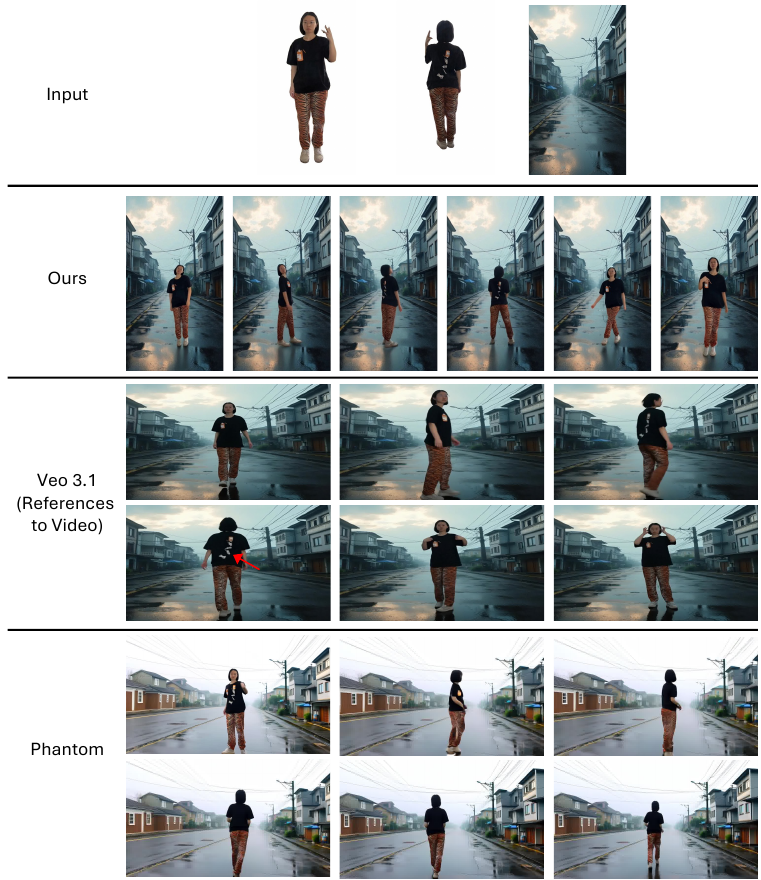}
    \caption{\textbf{Qualitative Comparison with Methods Trained on Other Datasets.} The first row shows the input: front selfie, back selfie, and background image. The remaining rows present the outputs from different methods. Both Veo 3.1 (References to Video) and Phantom take these three images as input along with a text description of motion. 
    Also, they only support landscape-mode video generation.  
    In addition, Phantom fails to effectively utilize the input background image, leading to inconsistent scene composition.
    }
    \label{fig:comparison_veo1}
\end{figure*}

\begin{figure*}[!t]
    \centering
\includegraphics[width=0.95\linewidth]{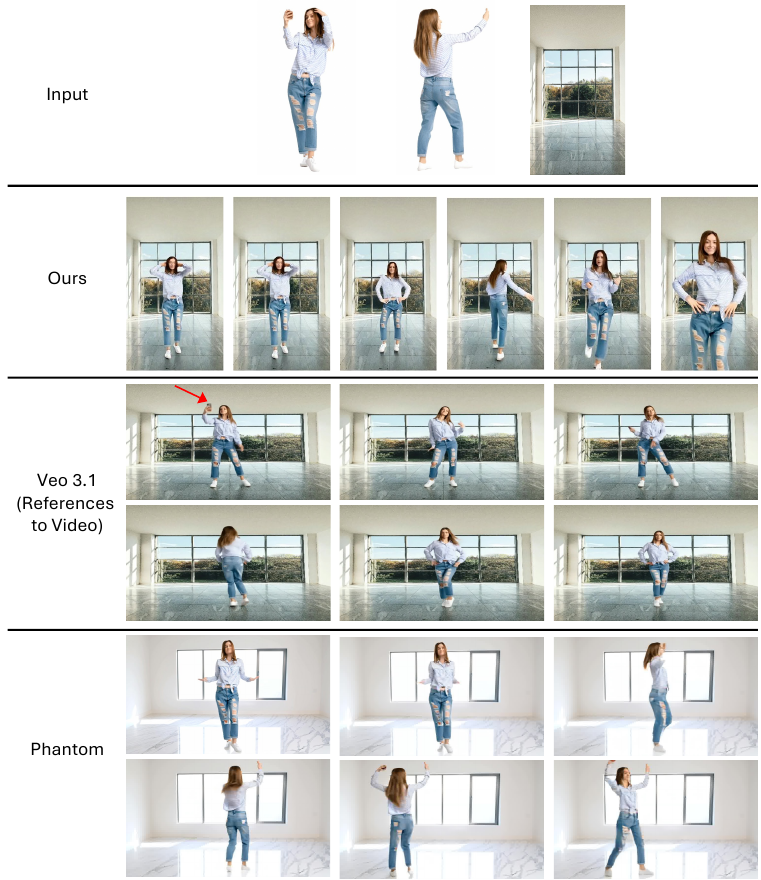}
    \caption{\textbf{Qualitative Comparison with Methods Trained on Other Datasets.} The first row shows the input: front selfie, back selfie, and background image. The remaining rows present the outputs from different methods. Both Veo 3.1 (References to Video) and Phantom take these three images as input along with a text description of motion. 
    Also, they only support landscape-mode video generation.  
    In addition, Phantom fails to effectively utilize the input background image, leading to inconsistent scene composition.
    }
    \label{fig:comparison_veo2}
\end{figure*}

\begin{figure*}[!t]
    \centering
\includegraphics[width=0.95\linewidth]{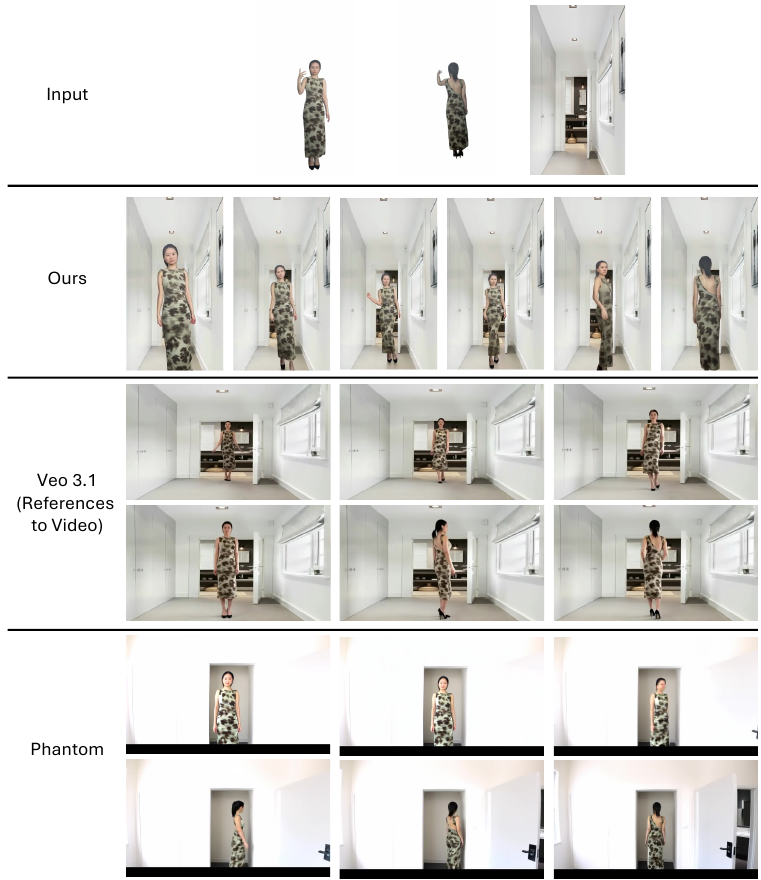}
    \caption{\textbf{Qualitative Comparison with Methods Trained on Other Datasets.} The first row shows the input: front selfie, back selfie, and background image. The remaining rows present the outputs from different methods. Both Veo 3.1 (References to Video) and Phantom take these three images as input along with a text description of motion. 
    Also, they only support landscape-mode video generation.  
    In addition, Phantom fails to effectively utilize the input background image, leading to inconsistent scene composition.
    }
    \label{fig:comparison_veo3}
\end{figure*}

Fig.~\ref{fig:comparison_veo1}, Fig.~\ref{fig:comparison_veo2}, and Fig.~\ref{fig:comparison_veo3} show qualitative comparisons. 
Both Veo 3.1 (References to Video) and Phantom only support landscape-mode video generation. 
Phantom generates inaccurate outfit (see Fig.~\ref{fig:comparison_veo1}). 
It also struggles to maintain consistency with the provided background image, often leading to incorrect scene composition. Veo 3.1 may produce inaccurate visual details (see the red arrow in Fig.~\ref{fig:comparison_veo1} and Fig.~\ref{fig:comparison_veo2}).


\subsection{Ablation Study}

\subsubsection{Qualitative Comparison}

\begin{figure*}[!t]
    \centering
\includegraphics[width=0.95\linewidth]{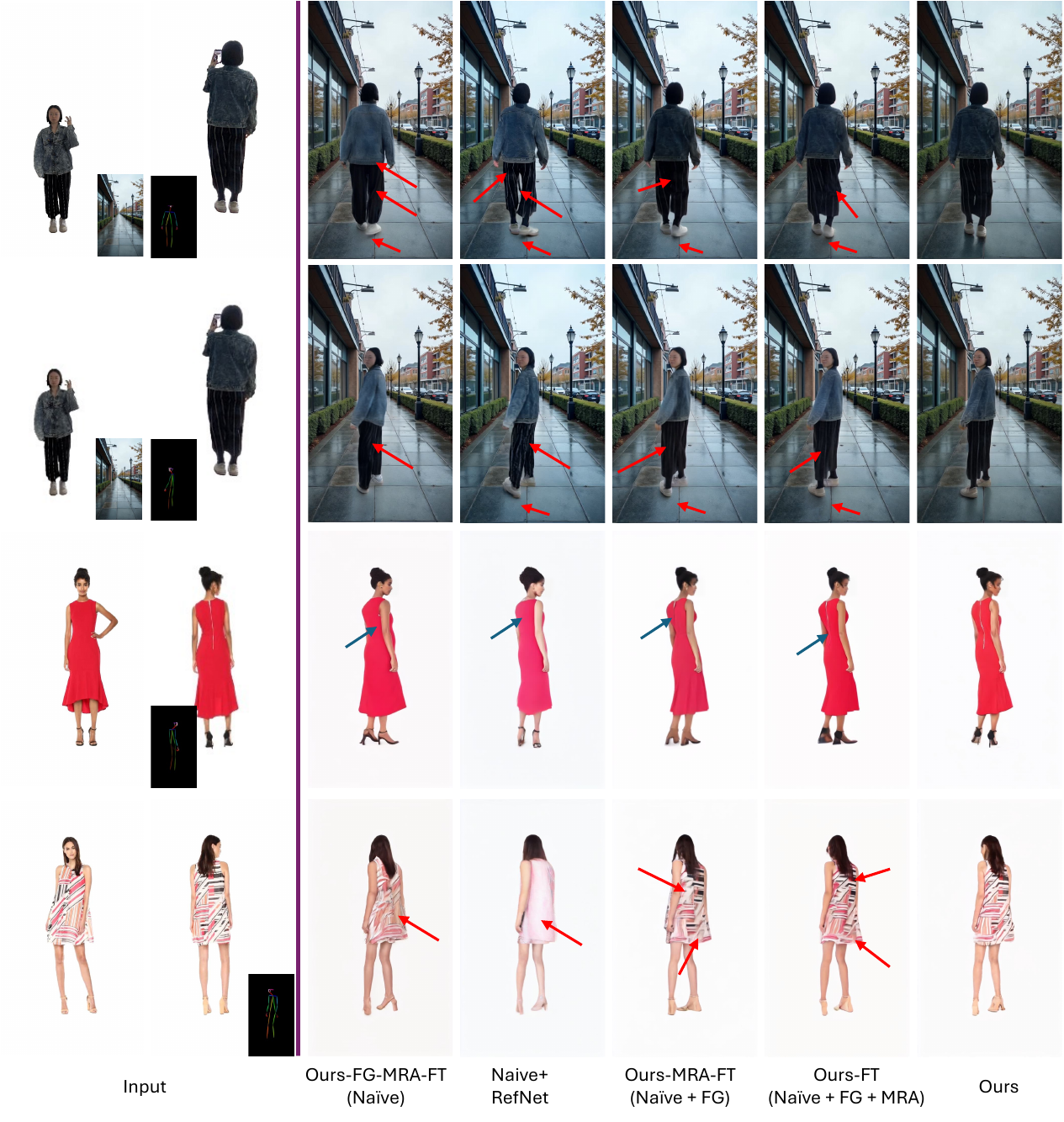}
\caption{\textbf{Model Ablations. }  The inputs are shown in the left two columns.
The right five columns showcase the generated results from various variants and our full model. 
All results are post-processed using face refinement.  
\textit{Ours-FG-MRA-FT (naive)} fails to render accurate back views.
\textit{Naive+RefNet} incorporates ReferenceNet, improving back views but creates artifacts (row 1 to 2) and fails in the case of a white background (row 3 to 4).
\textit{Ours-MRA-FT (Naïve + FG)} uses our frame generation strategy and produces better back views than naive methods.
\textit{Ours-FT (Naïve + FG + MRA)} additionally uses multi-reference attention and enhances back view patterns.
Importantly, all the variants fail to produce realistic reflections or generate weaker, blurry reflections on the ground (rows 1 and 2), whereas our method successfully achieves this.
In summary, our full model delivers sharper results with more accurate patterns, along with reasonable shadow and reflection generation.
}
    \label{fig:ablation_supp}
\end{figure*}

Fig.~\ref{fig:ablation_supp} shows additional comparisons of our variants with the full method. We observe the following:
(1) \textit{Ours-FG-MRA-FT (naïve)} fails to render accurate back views, indicating that simple modifications do not effectively encode features from additional reference images.
(2) \textit{Naïve+RefNet} incorporates ReferenceNet, improving back views but introducing artifacts (rows 1 to 2) and failing in the case of a white background (rows 3 to 4). This demonstrates that using ReferenceNet reduces the model’s generalization ability to unseen backgrounds. Additionally, it introduces extra parameters for training, which negatively impacts training efficiency.
(3) \textit{Ours-MRA-FT (Naïve + FG)} applies our frame generation strategy, producing better back views than naive methods, but still suffers from blurry patterns.
(4) \textit{Ours-FT (Naïve + FG + MRA)} further integrates multi-reference attention, enhancing back view patterns.

Importantly, all the variants fail to produce realistic reflections or generate weaker, blurry reflections on the ground (rows 1 and 2). In contrast, our method effectively achieves this by leveraging a fine-tuning strategy specifically designed to enhance shadows and reflections.
In summary, our full model produces sharper results with more accurate patterns while effectively generating realistic shadows and reflections.

\subsubsection{Quantitative Comparison}
Tab.~\ref{tab:test}, \ref{tab:ubc}, and \ref{tab:tiktok} present the quantitative results of our model compared to its variants across the three datasets. As discussed in the main paper, our method outperforms most variants across all metrics by employing a novel frame generation strategy, a multi-reference attention, and a fine-tuning approach, resulting in improved appearance fidelity and temporal consistency.

The ablation variant, \textit{Ours-RIA}, achieves results comparable to \textit{Ours} on this dataset because the input front and back images share the same background as the ground truth (GT) frames, making data augmentation less essential in this scenario.

\subsubsection{Inaccurate Segmentation}
Our method relies on pre-segmentation of the input selfies. Therefore, we evaluate its robustness to segmentation inaccuracies.
In practice, we found the adopted SAM2~\cite{ravi2024sam2} robust for segmenting mirror selfies. For study purposes, we dilated and eroded segmentation masks in self-captured dataset by 5\% and 10\%, feeding these inaccurately segmented selfies to our model. The resulting LPIPS scores – 0.383 (5\% dilation), 0.386 (10\% dilation), 0.384 (5\% erosion), and 0.388 (10\% erosion) – were only slightly worse than with accurate masks (0.381).

\subsection{Evaluation on Motion Retrieval}
We collect a test set by asking five participants to perform fit-check motions (\eg, walks, twirls) while using a phone to record IMU data. Simultaneously, we capture video recordings of these motions with an external camera.
From each video, we extract the corresponding SMPL sequence as ground truth (GT), which is later used for evaluation, constructing a dataset of 20 IMU inputs and GT SMPL sequences.

Since there are no existing IMU-to-motion retrieval methods, we compare our approach against TMR~\cite{petrovich23tmr}, a text-to-motion retrieval baseline. To provide input for this baseline, we manually annotate textual descriptions for the motions in our test set.
An example text annotation for a motion is:  ``A person walks away from the camera with their back facing it, then turns left by half a circle.''

We compare the top-k retrieved motions with the GT motion using a pretrained motion encoder~\cite{tevet2022human} to evaluate their similarity. This motion encoder takes as input a sequence of motion parameters in the HumanML3D format~\cite{Guo_2022_CVPR}, which can be obtained from an SMPL sequence (details in \cite{tevet2022human}), and outputs a motion embedding. 
For evaluation, we input both the retrieved and GT motions into the encoder and compute their similarity based on the dot product of their motion embeddings. Our method achieves a similarity score of 0.848 for $k=1$ and 0.716 for $k=5$, outperforming TMR’s 0.641 ($k=1$) and 0.522 ($k=5$).
These results demonstrate the effectiveness of our approach, highlighting that IMU-based retrieval provides more fine-grained motion guidance than text-based retrieval.
\clearpage
\clearpage

\clearpage
\clearpage

\clearpage

\clearpage
\clearpage

\clearpage

\clearpage

\clearpage

\clearpage

\clearpage

\clearpage

\clearpage


{
    \small
    \bibliographystyle{ieeenat_fullname}
    \bibliography{main}
}
